\documentclass[10pt,twocolumn,letterpaper]{article}

\usepackage[pagenumbers]{cvpr} 

\usepackage[utf8]{inputenc} 
\usepackage[T1]{fontenc}    
\usepackage[pagebackref=true,breaklinks=true,colorlinks,bookmarks=false]{hyperref}
\usepackage{url}            
\usepackage{booktabs}       
\usepackage{amsfonts}       
\usepackage{nicefrac}       
\usepackage{microtype}      
\usepackage{xcolor}         
\usepackage{wrapfig}
\usepackage{graphicx}
\usepackage{multirow}
\usepackage{tabularx}
\usepackage{tabulary}
\usepackage{colortbl}

\usepackage{xfrac}

\usepackage{xspace}

\usepackage{listings}
\usepackage{upquote}  
\usepackage{xcolor}
\definecolor{codegreen}{rgb}{0,0.6,0}
\definecolor{codegray}{rgb}{0.5,0.5,0.5}
\definecolor{codepurple}{rgb}{0.58,0,0.82}
\definecolor{backcolour}{rgb}{0.95,0.95,0.92}

\lstdefinestyle{mystyle}{
    backgroundcolor=\color{backcolour},   
    commentstyle=\color{codegreen},
    keywordstyle=\color{magenta},
    numberstyle=\tiny\color{codegray},
    stringstyle=\color{codepurple},
    basicstyle=\ttfamily\footnotesize,
    breakatwhitespace=false,         
    breaklines=true,                 
    captionpos=b,                    
    keepspaces=true,                 
    numbers=left,                    
    numbersep=5pt,                  
    showspaces=false,                
    showstringspaces=false,
    showtabs=false,                  
    tabsize=2
}

\definecolor{todocolor}{rgb}{0.82, 0.41, 0.12}

\title{Scaling Vision Transformers}

\author{%
  \centerline{Xiaohua Zhai$^{\star}$, Alexander Kolesnikov$^{\star}$, Neil Houlsby, Lucas Beyer$^{\star}$} \vspace{2mm}\\
  \centerline{Google Research, Brain Team, Zürich} \vspace{1.5mm} \\
  \centerline{\texttt{{\{xzhai, akolesnikov, neilhoulsby, lbeyer\}}@google.com}}
}

\def\cvprPaperID{10350} 
\def\confName{CVPR}
\def\confYear{2022}

\begin{document}

\maketitle
{\let\thefootnote\relax\footnote{
{$^{\star}$equal contribution}}}

\begin{abstract}

Attention-based neural networks such as the Vision Transformer (ViT) have recently attained state-of-the-art results on many computer vision benchmarks. Scale is a primary ingredient in attaining excellent results, therefore, understanding a model's scaling properties is a key to designing future generations effectively. While the laws for scaling Transformer language models have been studied, it is unknown how Vision Transformers scale. To address this, we scale ViT models and data, both up and down, and characterize the relationships between error rate, data, and compute. Along the way, we refine the architecture and training of ViT, reducing memory consumption and increasing accuracy of the resulting models. As a result, we successfully train a ViT model with two billion parameters, which attains a new state-of-the-art on ImageNet of $90.45\%$ top-1 accuracy. The model also performs well for few-shot transfer, for example, reaching $84.86\%$ top-1 accuracy on ImageNet with only 10 examples per class.

\end{abstract}

\section{Introduction}\label{sec:intro}

\begin{figure}[t]
  \begin{center}
    \includegraphics[width=0.4\textwidth]{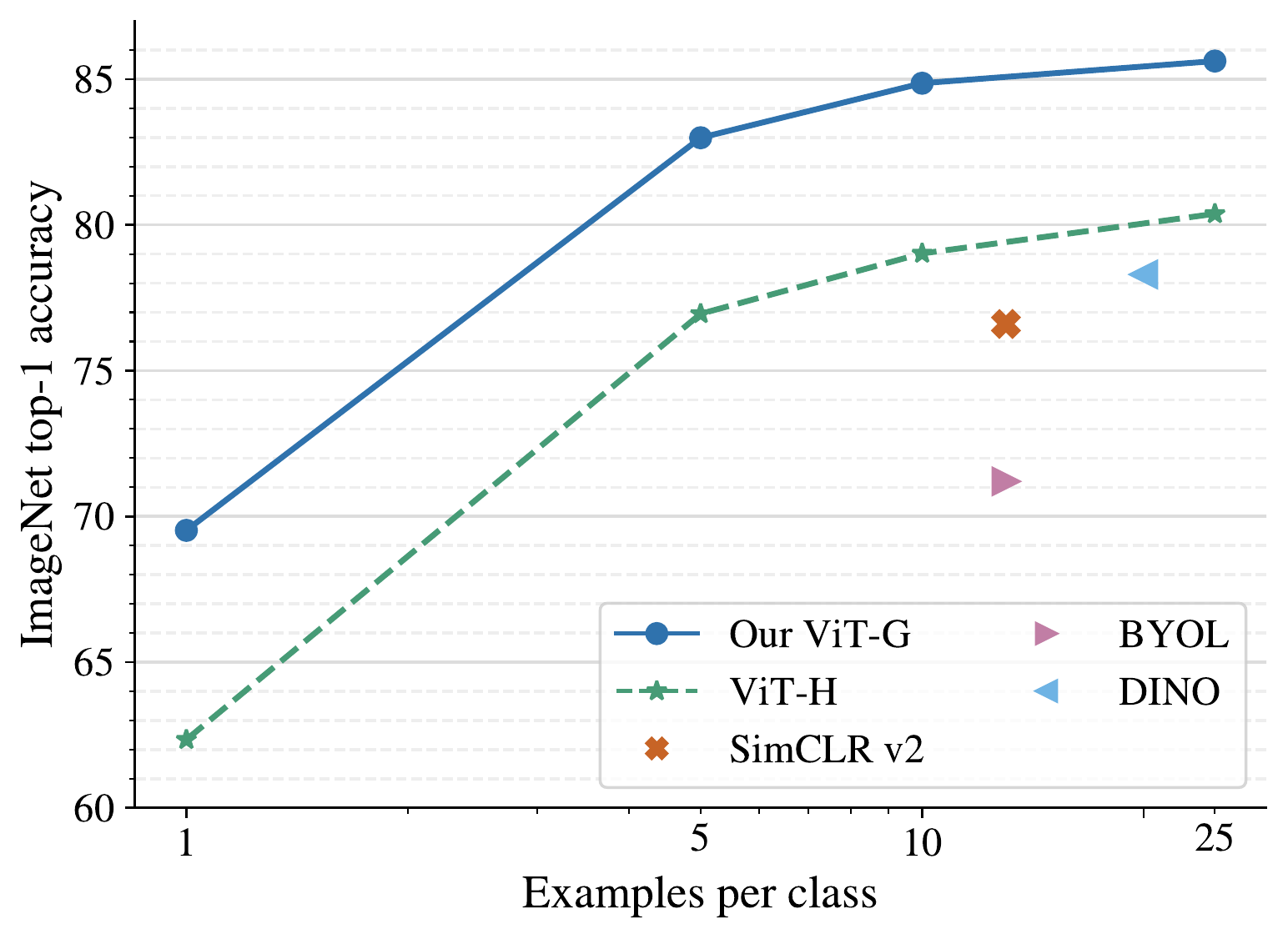}
  \end{center}
  \vspace{-1em}
  \caption{Few-shot transfer results. Our ViT-G model reaches 84.86\% top-1 accuracy on ImageNet with 10-shot linear evaluation.}\label{fig:few_shot}
  \vspace{-0.8em}
\end{figure}

Attention-based Transformer architectures~\cite{vaswani2017attention} have taken computer vision domain by storm~\cite{dosovitskiy2020, carion2020endtoend} and are becoming an increasingly popular choice in research and practice.  Previously, Transformers have been widely adopted in the natural language processing (NLP) domain~\cite{devlin2019bert, brown2020language}.  Optimal scaling of Transformers in NLP was carefully studied in~\cite{kaplan2020scaling}, with the main conclusion that large models not only perform better, but do use large computational budgets more efficiently. However, it remains unclear to what extent these findings transfer to the vision domain, which has several important differences. For example, the most successful pre-training schemes in vision are supervised, as opposed to unsupervised pre-training in the NLP domain.

In this paper we concentrate on scaling laws for transfer performance of ViT models pre-trained on image classification tasks. In particular, we experiment with models ranging from five million to two billion parameters, datasets ranging from one million to three billion training images and compute budgets ranging from below one TPUv3 core-day to beyond $10\,000$ core-days. Our main contribution is a characterization of the performance-compute frontier for ViT models, on two datasets.

\begin{figure*}[t]
  \begin{center}
  \vspace{-2em}
    \includegraphics[width=0.385\textwidth]{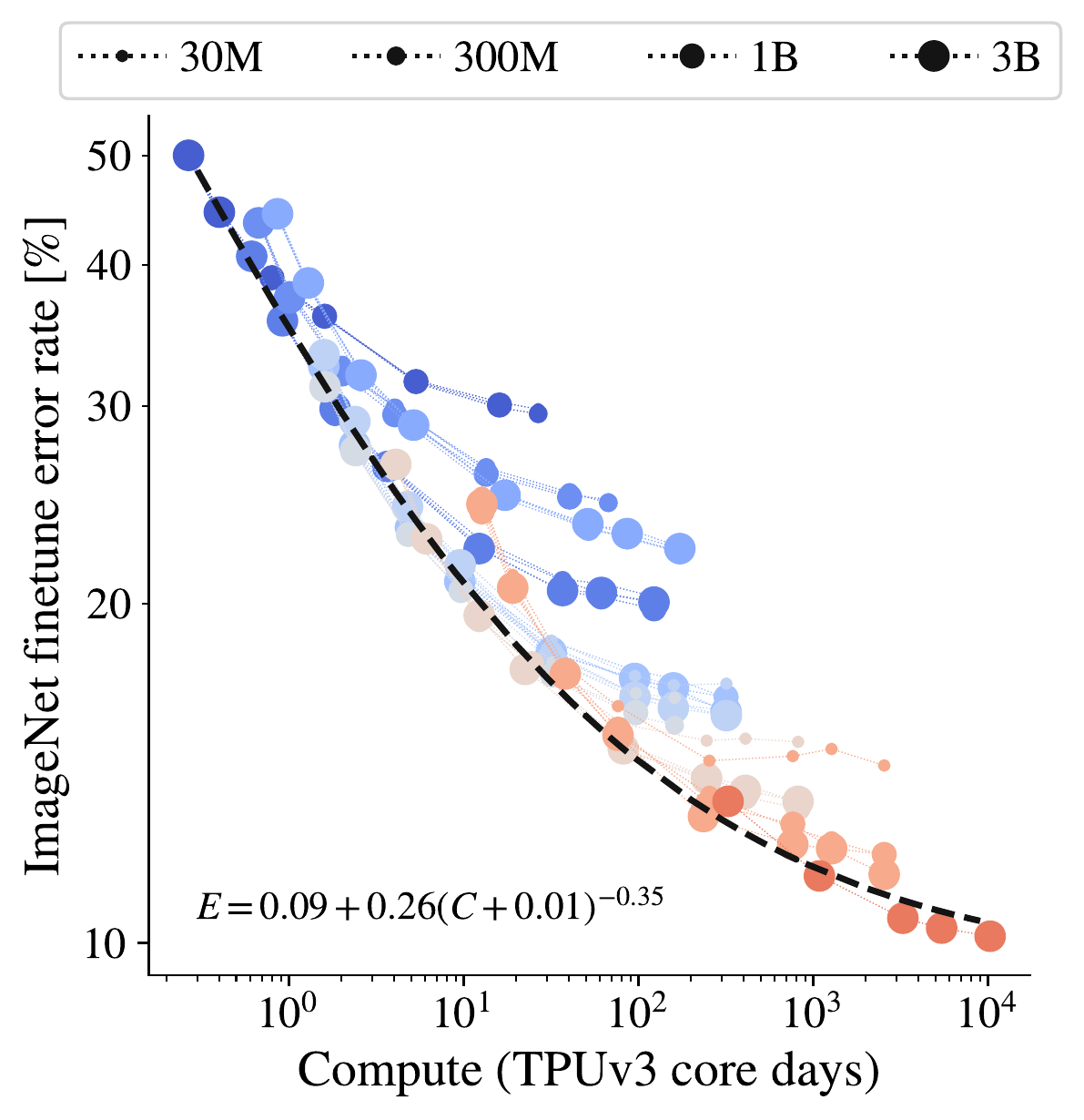}
    \includegraphics[width=0.605\textwidth]{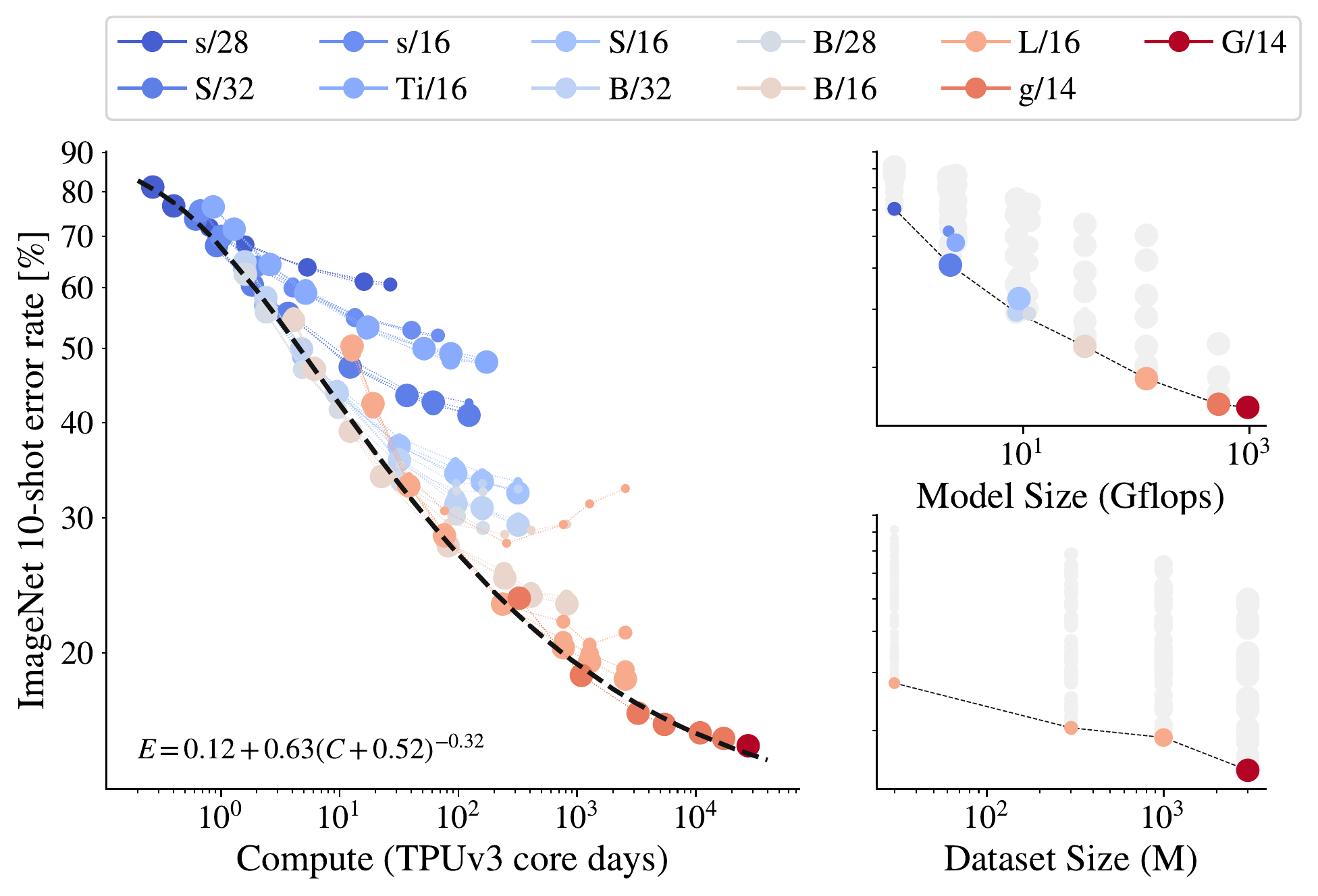}
  \end{center}
  \vspace{-1em}
  \caption{
  \textbf{Left}/\textbf{Center}: Representation quality, measured as ImageNet finetune and linear 10-shot error rate, as a function of total training compute.
  A saturating power-law approximates the Pareto frontier fairly accurately.
  Note that smaller models (blue shading), or models trained on fewer images (smaller markers), saturate and fall off the frontier when trained for longer.
  \textbf{Top right}: Representation quality when bottlenecked by model size.
  For each model size, a large dataset and amount of compute is used, so model capacity is the main bottleneck.
  Faintly-shaded markers depict sub-optimal runs of each model.
  \textbf{Bottom Right}: Representation quality by datasets size.
  For each dataset size, the model with an optimal size and amount of compute is highlighted, so dataset size is the main bottleneck.
  }\label{fig:teaser_pic}
  \vspace{-1em}
\end{figure*}

Along the way, we create an improved large-scale training recipe. We investigate training hyper-parameters and discover subtle choices that make drastic improvements in few-shot transfer performance. 
The few-shot transfer evaluation protocol has also been adopted by previous large-scale pre-training efforts in NLP domain~\cite{gpt3}.
Specifically, we discover that very strong L2 regularization, applied to the final linear prediction layer only, results in a learned visual representation that has very strong few-shot transfer capabilities.
For example, with just a single example per class on the ImageNet dataset (which has $1\,000$ classes), our best model achieves 69.52\% accuracy; and with \emph{10 examples per class it attains 84.86\%}.  In addition, we substantially reduce the memory footprint of the original ViT model proposed in~\cite{dosovitskiy2020}. We achieve this by hardware-specific architecture changes and a different optimizer. As a result, we train a model with two billion parameters and attain a new \emph{state-of-the-art 90.45\% accuracy on ImageNet}.

\section{Core Results}\label{sec:core}

We first present our main results on scaling trends, before presenting detailed architecture and training protocol improvements in Section~\ref{sec:setup}.
In the following experiments, we train several ViT models on both public ImageNet-21k~\cite{imagenet} dataset and privately gathered images, up to three billion weakly-labelled images.
We vary the architecture size, number of training images, and training duration.
All models are trained on TPUv3, thus total compute is measured in TPUv3 core-days.
To evaluate the quality of the representation learned by the models, we measure (i) few-shot transfer via training a linear classifier on frozen weights, (ii) transfer via fine-tuning the whole model on all data, both to multiple benchmark tasks.

\subsection{Scaling up compute, model and data together}\label{sec:scaling_up}

Figure~\ref{fig:teaser_pic} shows both the 10-shot linear evaluation and finetuning evaluation on ImageNet~\cite{imagenet}.
Similar trends on other datasets, Oxford IIIT Pets~\cite{pets}, CIFAR-100~\cite{cifar}, and Caltech-UCSD Birds~\cite{cub} are presented in the Appendix, Figure~\ref{fig:scaling_laws_all_results}.
For each combination of model size and data size we pre-train for various numbers of steps.
In Figure~\ref{fig:teaser_pic}, connected points represent the same model trained for a different number of steps. 
We make the following observations.

First, \textit{scaling up compute, model and data together improves representation quality}.
In the left plot and center plot, the lower right point shows the model with the largest size, dataset size and compute achieving the lowest error rate.
However, it appears that at the largest size the models starts to saturate, and fall behind the power law frontier (linear relationship on the log-log plot in Figure~\ref{fig:teaser_pic}).

Second, \textit{representation quality can be bottlenecked by model size}. 
The top-right plot shows the best attained performance for each model size. 
Due to limited capacity, small models are not able to benefit from either the largest dataset, or compute resources.
Figure~\ref{fig:teaser_pic}, left and center, show the Ti/16 model tending towards a high error rate, even when trained on a large number of images.

Third, \textit{large models benefit from additional data, even beyond 1B images}. 
When scaling up the model size, the representation quality can be limited by smaller datasets; even 30-300M images is not sufficient to saturate the largest models.
In Figure~\ref{fig:teaser_pic}, center, the error rate of L/16 model on the the 30M dataset does not improve past 27\%.
On the larger datasets, this model attains 19\%.
Further, when increasing the dataset size, we observe a performance boost with big models, but not small ones.
The largest models even obtain a performance improvement the training set size grows from 1B to 3B images (Figure~\ref{fig:teaser_pic}, bottom right).
For small models, however, such as Ti/16 or B/32, increasing the dataset size does not help.
For example, in Figure~\ref{fig:teaser_pic}, left and center, all of the curves for Ti/16 overlap, showing that this model achieves the same performance irrespective of the dataset size.

\begin{figure*}[t]
  \begin{center}
    \includegraphics[width=\linewidth]{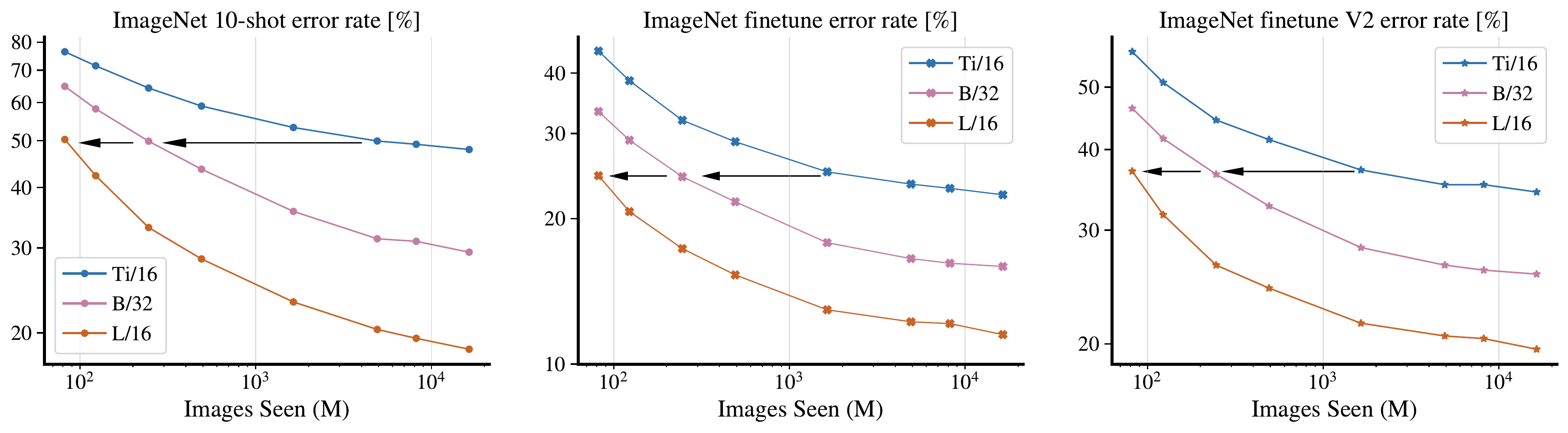}
  \end{center}
  \caption{Error rate on ImageNet, with respect to images seen during pre-training. Big models are more sample efficient, which is consistent across diverse setups: few-shot transfer on the frozen representations, fine-tune the network on ImageNet, and evaluate the fine-tuned models on the v2 test set.}
  \label{fig:sample_efficiency}
\end{figure*}

\subsection{Double-saturating power law}\label{sec:power_laws}

Figure~\ref{fig:teaser_pic}, left and center, show the Pareto frontier of representation quality versus training compute.
The frontier contains the models with the best allocation of compute to model shape and training duration.

For over two orders of magnitude of compute, the relationship between compute and performance follows a power-law ($E=aC^b$), resulting in a straight line on the log-log plot.
However, we observe ``saturation'' at both ends of the compute spectrum.
At the higher end of compute, the largest models do not tend towards zero error-rate.
If we extrapolate from our observations, an infinite capacity model will obtain a non-zero error.
This effect has also been observed for generative models~\cite{henighan2020scaling}.
The authors of~\cite{henighan2020scaling} refer to this residual error as the ``irreducible entropy'' of the task.
Since we plot error rate, the information-theoretic interpretation does not apply, but our observations support the notion of fundamental performance ceilings for ImageNet~\cite{beyer2020imagenet}.
In terms of the law, this saturation corresponds to an additive constant to the error rate: $c$ in $E=aC^{-b}+c$.

At the lower end of the compute spectrum, we see a saturation for smaller models;
the performance of the smallest model is better than that would be predicted by a power-law.
This saturation occurs because even trivial solutions can achieve non-zero error.
For example, predicting the majority class (almost zero compute) will achieve an accuracy related to its occurence frequency in the test set.
This lower bound is not observed in~\cite{henighan2020scaling}, either because their smallest model is large enough to avoid this region, or because log-loss saturates at worse performances than accuracy (it will saturate eventually).
This saturation corresponds to a shift in the x-axis: $d$ in $E=a(C+d)^{-b}+c$.
This constant indicates that the zero-compute model will still obtain non-zero accuracy.

\subsection{Big models are more sample efficient} \label{sec:sample_efficiency}

Figure~\ref{fig:sample_efficiency} shows the representation quality with respect to the total number of images ``seen'' (batch size times number of steps) during pre-training. 
In addition to ImageNet fine-tuning and linear 10-shot results on the public validation set, we also report results of the ImageNet fine-tuned model on the ImageNet-v2 test set~\cite{recht2019imagenet} as an indicator of robust generalization.
Three ViT models pre-trained on three billion images are presented in this plot. 

We observe that \textit{bigger models are more sample efficient}, reaching the same level of error rate with fewer seen images.
For 10-shot, the Ti/16 model needs to see nearly 100 times more images to match the representation quality of the L/16 model.
When fine-tuning, this factor reduces from 100 to about 20.
Our results suggest that with sufficient data, training a larger model for fewer steps is preferable.
This observation mirrors results in language modelling and machine translation~\cite{kaplan2020scaling,gshard}.

\begin{figure}[b]
  \begin{center}
    \includegraphics[width=0.48\textwidth]{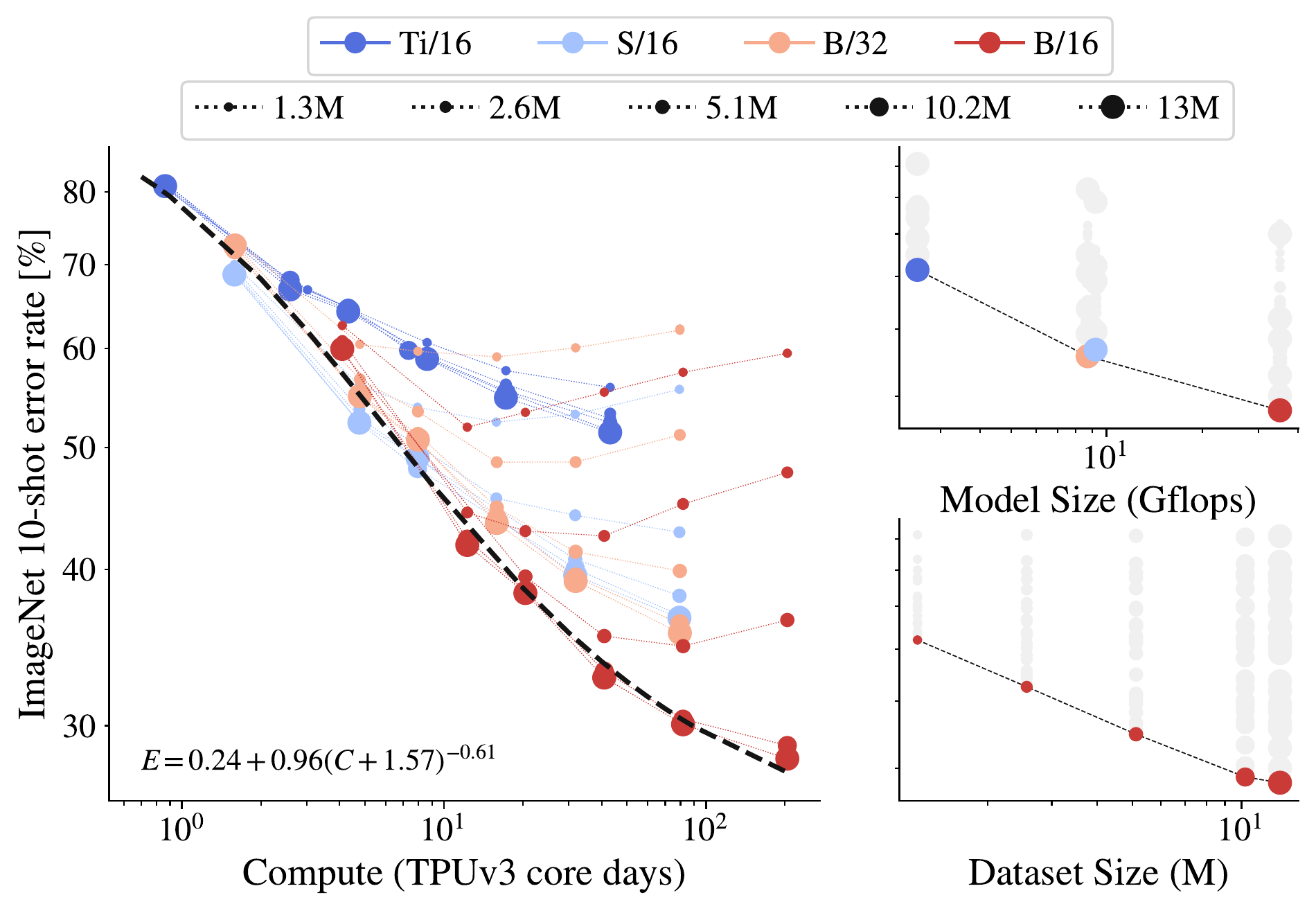}
  \end{center}
  \caption{Results on the ImageNet-21k dataset. \textbf{Left}: Representation quality, measured as ImageNet linear 10-shot error rate, as a function of total training compute. The double-saturating power law still applies. \textbf{Right}: Representation quality by model sizes and dataset sizes.}\label{fig:scaling_laws_i21k}
\end{figure}

\begin{table*}[t]
  \newcolumntype{C}{>{\centering\arraybackslash}X}
  \newcolumntype{R}{>{\raggedleft\arraybackslash}X}
  \setlength{\tabcolsep}{0pt}
  \setlength{\extrarowheight}{5pt}
  \renewcommand{\arraystretch}{0.75}
  \centering
  \caption{The results for ViT-G/14, compared to the previous state-of-the-art models.}\label{table:sota}
  \begin{tabularx}{\linewidth}{p{3.2cm}p{0.1cm}Cp{0.1cm}Cp{0.1cm}Cp{0.1cm}Cp{0.1cm}C}
    \toprule[1pt]
     \bf{Benchmark} && \bf{ImageNet} && \bf{INet V2} && \bf{INet ReaL} && \bf{ObjectNet} && \bf{VTAB (light)} \\
    \midrule
     NS (Eff.-L2)~\cite{xie2019selftraining} && 88.3 && 80.2 && - && 68.5 && - \\
     MPL (Eff.-L2)~\cite{pham2020meta} && 90.2 && - && \textbf{91.02} && - && - \\
     CLIP (ViT-L/14)~\cite{radford2021learning} && 85.4 && 75.9 && - && \textbf{72.3} && -\\
     ALIGN (Eff.-L2)~\cite{jia2021scaling} && 88.6 && 70.1 && - && - && -\\
     BiT-L (ResNet)~\cite{kolesnikov2019big} && 87.54 && - && 90.54 && 58.7 && 76.29\\
     ViT-H/14~\cite{dosovitskiy2020} && 88.55 && - && 90.72 && - && 77.63 \\
     \midrule
     Our ViT-G/14 && \textbf{90.45$\pm$0.03} && \textbf{83.33$\pm$0.03}  && 90.81$\pm$0.01 && 70.53$\pm$0.52 && \textbf{78.29$\pm$0.53} \\ 
    \bottomrule[1pt]
  \end{tabularx}
\end{table*}

\subsection{Do scaling laws still apply on fewer images?}\label{sec:public}

We extend the study to much fewer images, ranging from one million to 13 millions on the public ImageNet-21k dataset. 
In Figure~\ref{fig:scaling_laws_i21k} left, we found that the double-saturation power law \textit{still applies}, when varying model sizes, dataset sizes and compute resources. 
This indicates that the conclusions from the study generalizes well, and can guide future design choices for vision transformer architectures.
In Figure~\ref{fig:scaling_laws_i21k} right, we observe similar behaviors that the model performance are bottlenecked by the dataset size. 
When scaling up compute, model and data together, one gets the best representation quality.

\subsection{ViT-G/14 results}\label{sec:results}

We trained a large Vision Transformer, ViT-G/14, which contains nearly two billion parameters.
Section~\ref{sec:shape} details the architecture's shape.
We evaluate the ViT-G/14 model on a range of downstream tasks, and compare it to recent state-of-the-art results.
We fine-tune on ImaegNet, and report ImageNet~\cite{ILSVRC15}, ImageNet-v2~\cite{recht2019imagenet}, ReaL~\cite{beyer2020imagenet}, and ObjectNet~\cite{Barbu2019ObjectNetAL} accuracies. 
In addition, we report transfer learning result on the VTAB-1k benchmark consisting of 19 tasks~\cite{zhai2019largescale}. 

Figure~\ref{fig:few_shot} shows the few-shot transfer results on ImageNet.
ViT-G/14 outperforms the previous best ViT-H/14 model~\cite{dosovitskiy2020} by a large margin (more than 5\%), attaining \emph{84.86\% accuracy with 10 examples per class}.
Ten images per class is less than 1\% of ImageNet data (13 examples per class), as commonly used in self-supervised and semi-supervised learning~\cite{zhai2019s4l}.
For reference, Figure~\ref{fig:few_shot} shows three state-of-the-art self-supervised learning models, SimCLR v2~\cite{chen2020big} and BYOL~\cite{grill2020bootstrap}, using 1\% of ImageNet data, DINO~\cite{dino} using 20 examples per class.
Note, however, that these approaches are quite different:
ViT-G/14 uses large source of weakly-supervised data, and is pre-trained only once and transferred to different tasks.
Meanwhile, the self-supervised learning models use unlabeled but in-domain data for pre-training, and target a single task.

Table~\ref{table:sota} shows the results on the remaining benchmarks.
ViT-G/14 achieves \emph{90.45\% top-1 accuracy on ImageNet}, setting the new state-of-the art. 
On ImageNet-v2, ViT-G/14 improves 3\% over the Noisy Student model~\cite{xie2019selftraining} based on EfficientNet-L2. 
For ReaL, ViT-G/14 outperforms ViT-H~\cite{dosovitskiy2020} and BiT-L~\cite{kolesnikov2019big} by only a small margin, which indicates again that the ImageNet classification task is likely reaching its saturation point.
For ObjectNet, ViT-G/14 outperforms BiT-L~\cite{kolesnikov2019big} by a large margin, and is 2\% better than Noisy Student, but is about 2\% behind CLIP~\cite{radford2021learning}. 
Note that, unlike the other methods, CLIP does not fine-tune on ImageNet, and evaluates directly on ObjectNet, this likely improves its robustness.
Finally, when transferring the ViT-G/14 model to VTAB, it gets consistently better results with just a single hyper parameter across all tasks.
The state-of-the-art on VTAB using a heavyweight per-task hyperparameter sweep is 79.99~\cite{jia2021scaling}, we leave running a heavy sweep with ViT-G/14 to future work.

\section{Method details}\label{sec:setup}

We present a number of improvements to the ViT model and training.
These improvements are mostly simple to implement, and can significantly improve memory-utilization and model quality.
They allow us to train ViT-G/14 using data-parallelism alone, with the entire model fitting on a single TPUv3 core.

\subsection{Decoupled weight decay for the ``head''}

\begin{figure*}%
    \centering
    \raisebox{-0.47\height}{\includegraphics[width=0.65\textwidth]{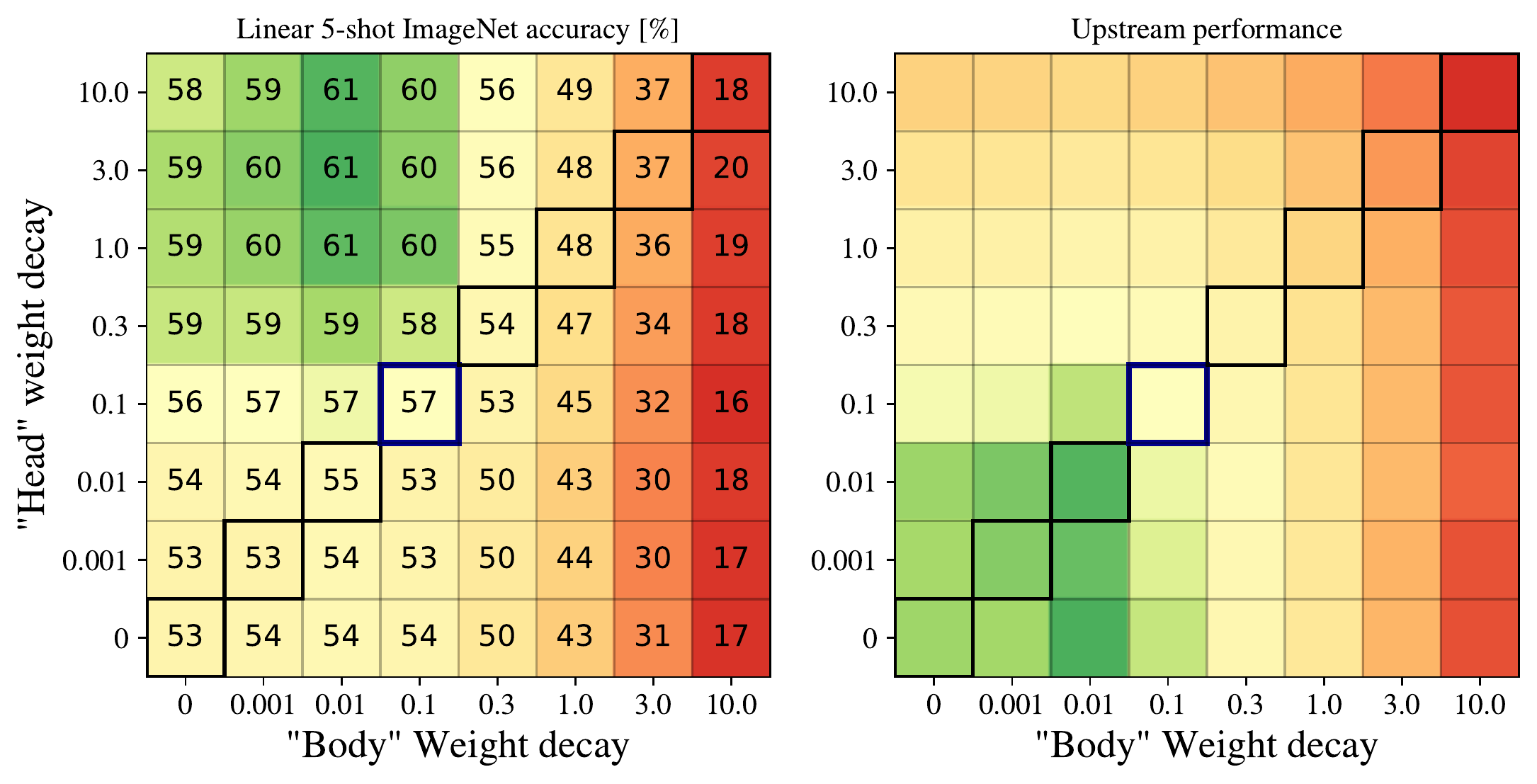}}%
    \hspace*{.1in}
    \raisebox{-0.5\height}{\includegraphics[width=0.34\textwidth]{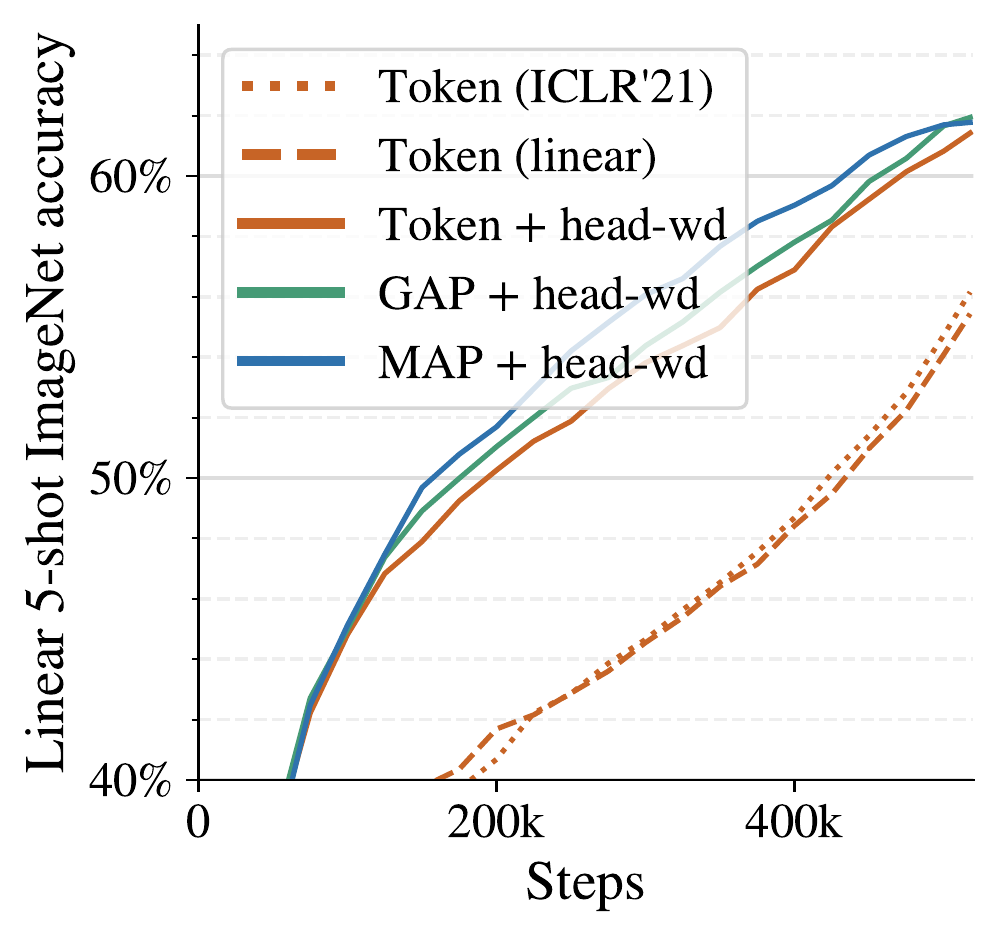}}%
    \caption{\textbf{Left and middle}: The dependence of 5-shot ImageNet accuracy and upstream performance depends on the weight decay strength. Normally, a single weight decay value is applied to all weights (corresponds to the diagonal on the heatmaps). We show that by using weight decay values for the ``head'' and the rest of the weights one significantly improves few-shot transfer performance.
    \textbf{Right}: Few-shot performance on ImageNet for different types of head.  A high weight decay on the head works equally well for all of them.}%
    \vspace{-1em}
    \label{fig:wd}%
\end{figure*}

Weight decay has a drastic effect on model adaptation in the low-data regime. We conduct an study of this phenomena at a mid-size scale.

We find that one can benefit from decoupling weight decay strength for the final linear layer (``head''), and for the remaining weights (``body'') in the model. 
Figure~\ref{fig:wd} demonstrates this effect: we train a collection ViT-B/32 models on JFT-300M, each cell corresponds to the performance of  different head/body weight decay values. 
The diagonal corresponds to using the same value for both decays.
One can observe that the best performance appears off-diagonal (i.e. with a decoupled weight decay for the head and body).
Interestingly, we observe that high weight decay in the head decreases performance on the pre-training (upstream) task (not shown), despite improving transfer performance.

We do not have a complete explanation of this phenomena. However, we hypothesize that a stronger weight decay in the head results in representations with larger margin between classes, and thus better few-shot adaptation.
This is similar to the main idea behind SVMs~\cite{cortes1995support}.
This large decay makes it harder to get high accuracy during upstream pre-training, but our main goal is high quality transfer.

\subsection{Saving memory by removing {\tt [class]} token}\label{sec:head}

The largest VIT model from~\cite{dosovitskiy2020} uses $14 \times 14$ patches with $224 \times 224$ images. This results in $256$ visual ``tokens'', where each one corresponds to an image patch. On top of this, ViT models have an extra {\tt [class]} token, which is used to produce the final representation, bringing the total number of tokens to $257$.

For ViT models, current TPU hardware pads the token dimension to a multiple of $128$, which may result in up to a $50\%$ memory overhead.
To overcome this issue we investigate alternatives to using the extra {\tt [class]} token.
In particular, we evaluate global average pooling (GAP) and multihead attention pooling (MAP)~\cite{lee2019set} to aggregate representation from all patch tokens.
We set the number of heads in MAP to be equal to the number of attention heads in the rest of the model.
To further simplify the head design we remove final non-linear projection before the final prediction layer, which was present in the original ViT paper.

To choose the best head, we perform a side-by-side comparison of a {\tt [class]} token and GAP/MAP heads.
Results are summarized in Figure~\ref{fig:wd}~(right).
We find that all heads perform similarly, while GAP and MAP are much more memory efficient due to the aforementioned padding considerations. We also observe that non-linear projection can be safely removed.
Thus, we opt for the MAP head, since it is the most expressive and results in the most uniform architecture. 
MAP head has also been explored in~\cite{cait}, in a different context for better quality rather than saving memory. 

\begin{figure*}[t]
  \begin{center}
    \includegraphics[width=0.8\linewidth]{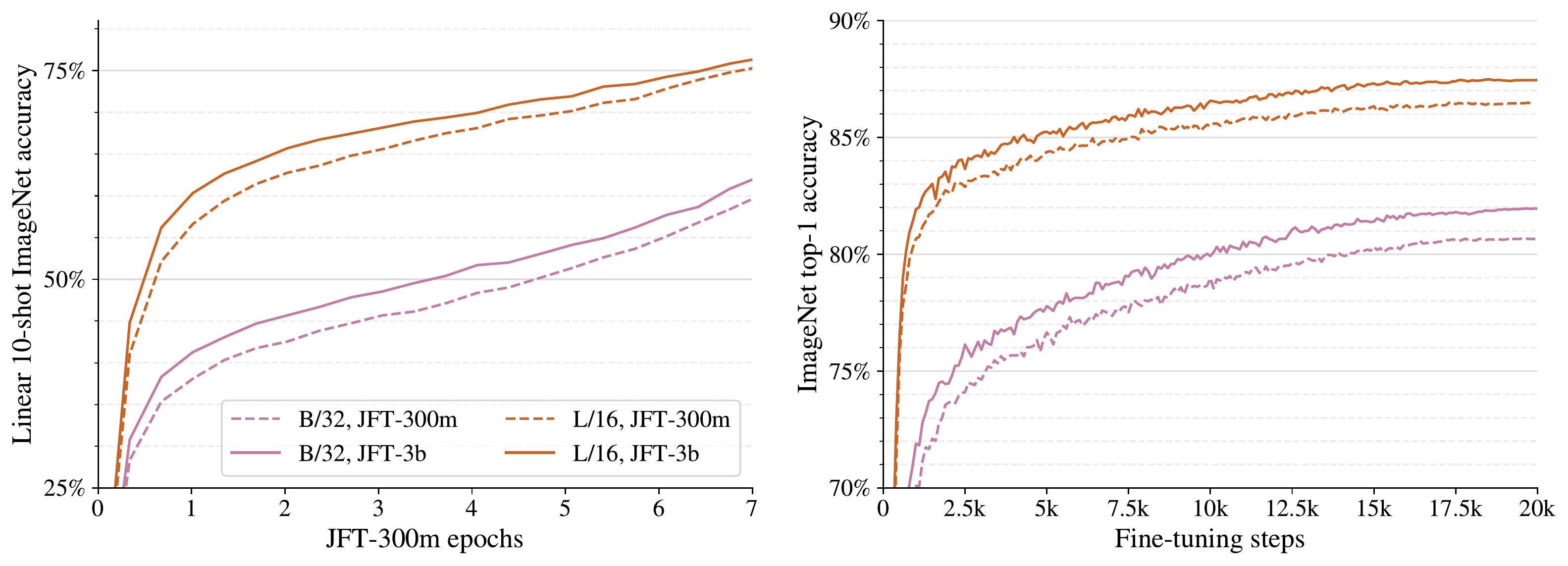}
  \end{center}
  \vspace{-0.5em}
  \caption{The effect of switching from JFT-300M to JFT-3B, without any further scaling. Both small and large models benefit from this change, by an approximately constant factor, both for linear few-shot evaluation (\textbf{left}) and transfer using the full dataset (\textbf{right}).}
  \label{fig:jft_300m_3b}
\end{figure*}

\subsection{Scaling up data}\label{sec:dataset}

For this study, we use the proprietary JFT-3B dataset, a larger version of the JFT-300M dataset used in many previous works on large-scale computer vision models~\cite{sun2017unreasonable,kolesnikov2019big,dosovitskiy2020}.
This dataset consists of nearly 3 billion images, annotated with a class-hierarchy of around 30k labels via a semi-automatic pipeline.
Thus, the data and associated labels are noisy.
We ignore the hierarchical aspect of the labels and use only the assigned labels as targets for multi-label classification via a sigmoid cross-entropy loss, following~\cite{kolesnikov2019big,dosovitskiy2020}.

We have conducted sensitive category association analysis as described in~\cite{aka2021measuring}. We measured (per label) the distribution of sensitive categories across the raw data, the cleaned data, the models trained on this data, and labels that were verified by human raters. Human raters additionally assisted in removing offensive content from the dataset.

Figure~\ref{fig:jft_300m_3b} shows an ablation of the effect of changing from JFT-300M to JFT-3B on model performance, even when scale is not increased.
Figure~\ref{fig:jft_300m_3b}, left shows linear 10-shot ImageNet performance evaluated throughout.
We observe that JFT-3B  results in a better model, even before the model has completely one epoch of JFT-300M.
Therefore, overfitting JFT-300M is not the sole cause of the improvement.
This difference can be seen even for the small B/32 model as well as the larger L/16.
We fine-tune the models to the full ImageNet dataset (right), and confirm that these improvements transfer to a full fine-tuning setup.
Overall, the change in dataset improves transfer to ImageNet by about 1\% for both small and large models. 
Other than the performance improvement, training behavior is similar on JFT-300M and JFT-3B.
Most importantly, JFT-3B allows us to scale up further with fewer concerns about overfitting and regularization.

\textbf{Deduplication.} We remove all images from the JFT-3B dataset that are near-duplicates of images from both train set and test set of datasets we evaluate on. Overall we identified and removed 927k duplicate images from JFT-3B.

\subsection{Memory-efficient optimizers}\label{sec:optims}

When training large models, storage required for model parameters becomes a bottleneck. Our largest model, ViT-G, has roughly two billion parameters, which occupies 8 GiB of device memory. To make things much worse, the Adam optimizer that is commonly used for training Transformers, stores two additional floating point scalars per each parameter, which results in an additional two-fold overhead (extra 16 GiB). To tackle the overhead introduced by the Adam optimizer we explore two modifications.

\textbf{Adam with half-precision momentum}. We empirically observe that storing momentum in half-precision (\texttt{bfloat16} type) does not affect training dynamics and has no effect on the outcome. This allows to reduce optimizer overhead from 2-fold to 1.5-fold. Notably, storing the second momentum using half-precision resulted in a significant performance deterioration.

\textbf{Adafactor optimizer.} The above optimizer still induces a large memory overhead. Thus, we turn our attention to the Adafactor optimizer~\cite{adafactor}, which stores second momentum using rank 1 factorization. From practical point of view, this results in the negligible memory overhead. However, the Adafactor optimizer did not work out of the box, so we make the following modifications:
\begin{itemize}
    \item We re-introduce the first momentum in half-precision, whereas the recommended setting does not use the first momentum at all.
    \item We disable scaling of learning rate relative to weight norms, a feature that is part of Adafactor.
    \item Adafactor gradually increases the second momentum from $0.0$ to $1.0$ throughout the course of training. In our preliminary experiments, we found that clipping the second momentum at $0.999$ (Adam's default value) results in better convergence, so we adopt it.
\end{itemize}
The resulting optimizer introduces only a 50\% memory overhead on top the space needed to store model's parameters. 

We observe that both proposed optimizers perform on par with  or slightly better than the original Adam optimizer.
We are aware of other memory-efficient optimizers~\cite{zero_optimizer,adam_1bit}, we leave the exploration to future work.

\subsection{Learning-rate schedule}

In our study we want to train each of the models for several different durations in order to measure the trade-off between model size and training duration.
When using linear decay, as in~\cite{dosovitskiy2020}, each training duration requires its own training run starting from scratch, which would be an inefficient protocol.

\begin{figure}[t]
  \begin{center}
  \vspace{-0.5em}
    \includegraphics[width=0.45\textwidth]{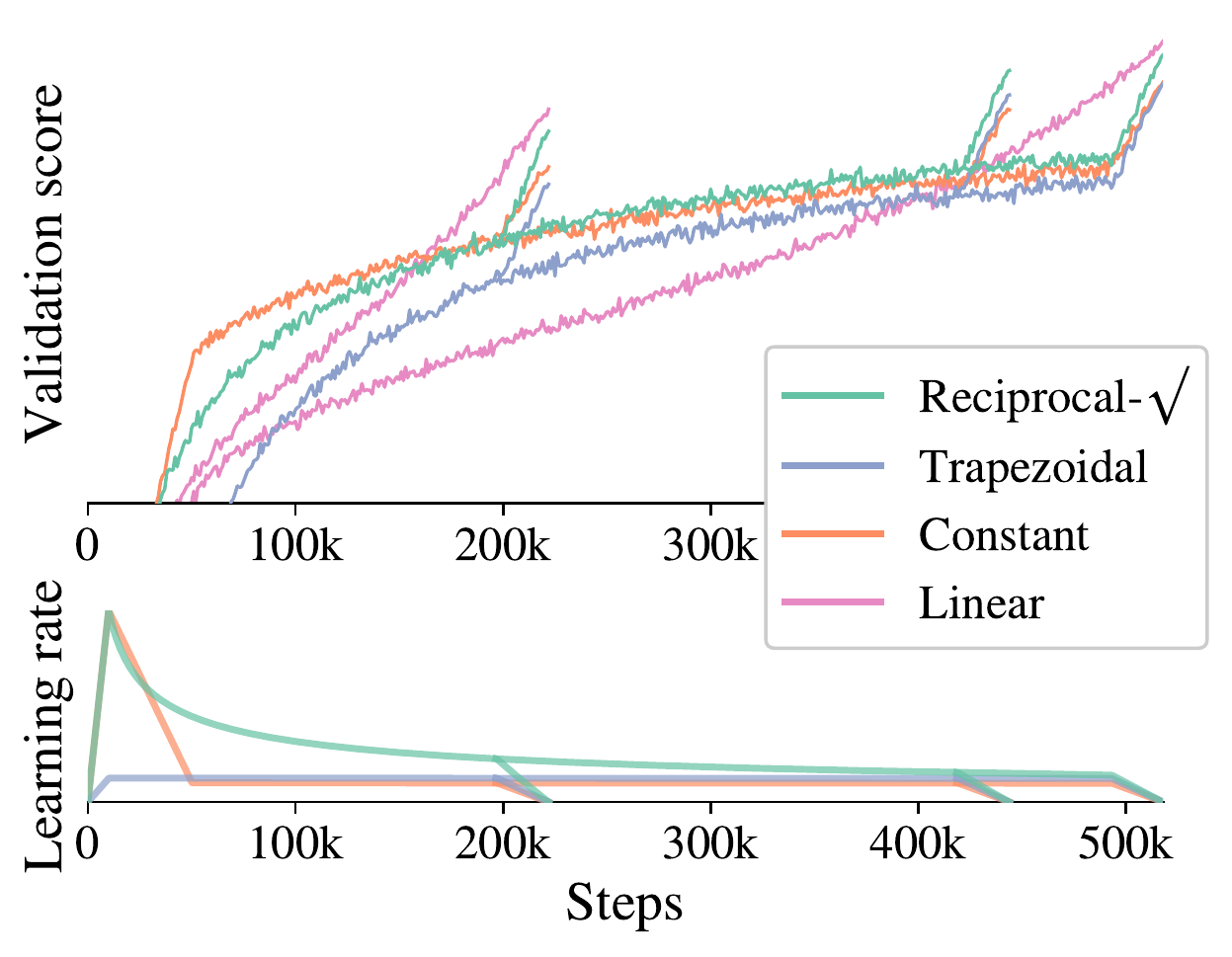}
  \end{center}
  \vspace{-1.2em}
  \caption{Various ``infinite'' learning-rate schedules, along with the finite linear one for reference.}\label{fig:schedules}
\end{figure}

Inspired by~\cite{instagram}, we address this issue by exploring learning-rate schedules that, similar to the \emph{warmup} phase in the beginning, include a \emph{cooldown} phase at the end of training, where the learning-rate is linearly annealed toward zero.
Between the warmup and the cooldown phases, the learning-rate should not decay too quickly to zero.
This can be achieved by using either a constant, or a reciprocal square-root schedule for the main part of training.
Figure~\ref{fig:schedules}~(bottom) depicts several of these options, with a cooldown after approximately 200\,k, 400\,k, and 500\,k steps.
The upper half of Figure~\ref{fig:schedules} shows the validation score (higher is better) for each of these options and their cooldowns, together with two linear schedules for reference.
While the linear schedule is still preferable when one knows the training duration in advance and does not intend to train any longer, all three alternatives come reasonably close, with the advantage of allowing indefinite training \emph{and} evaluating multiple training durations from just one run.
For each of the schedules, we optimized the learning-rate and the exact shape.
We have also briefly tried cyclic learning-rate schedules, however they seemed to perform much worse and we have not investigated further.
We therefore opt for the reciprocal square-root schedule.

\subsection{Selecting model dimensions}\label{sec:shape}

\begin{figure*}[t]
  \begin{center}
    \includegraphics[width=0.92\linewidth]{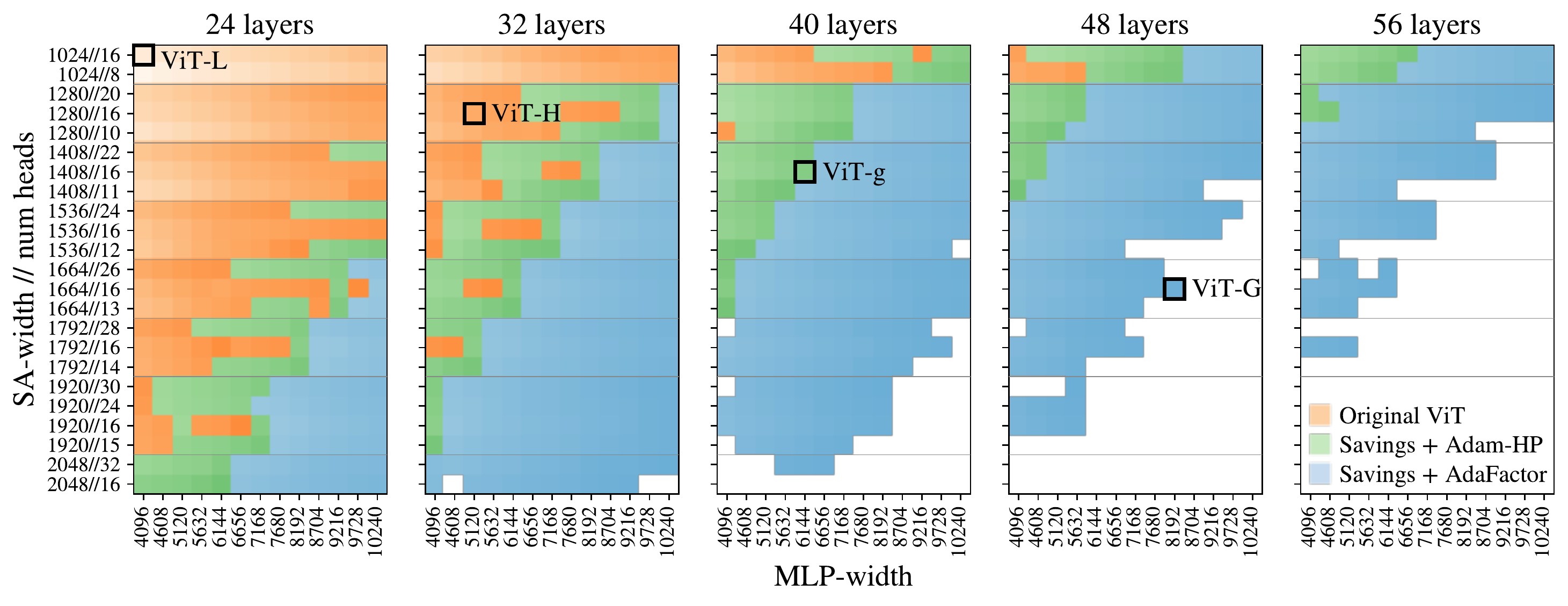}
  \end{center}
  \vspace{-1.5em}
  \caption{Combined results of the ``Shapefinder'' simulation for the original ViT in orange, our improvements together with half-precision Adam (\eg ViT-g) in green, and finally with our modified AdaFactor in blue. White areas ran out of memory. The brightness of the dot corresponds to its relative training speed.}
  \label{fig:shapefinder}
\end{figure*}

ViT models have many parameters that control the model's shape, and we refer to the original publication for full details.
Briefly, these include the \emph{patch-size}, the number of encoder blocks (\emph{depth}), the dimensionality of patch embeddings and self-attention (\emph{width}), the number of attention \emph{heads}, and the hidden dimension of MLP blocks (\emph{MLP-width}).
On top of this, we rely on the XLA compiler to optimize our models for runtime speed and memory footprint.
Behind the scenes, XLA uses complex heuristics to compile a model into code for a specific hardware that trades off memory and speed optimally.
As a result, it is hard to predict which model configurations will fit into memory on a single device.

Therefore we run an extensive simulation, where we instantiate a large amount of ViTs of various shapes, and attempt to train them for a few steps, without considering the quality.
We vary the depth, width, heads, and MLP-width, but keep the patch-size at 14\,px. 
In this way, we measure their speed and whether or not a given model fits into the device's memory.
Figure~\ref{fig:shapefinder} summarizes the result of this simulation.
Each block corresponds to one model configuration, the shade of the block corresponds to its training speed (brighter is faster). 
Orange blocks show which original ViT models, without any of our modifications, fit.
Green blocks then further include the memory savings described in Section~\ref{sec:head} coupled with the half-precision Adam described in Section~\ref{sec:optims}.
Finally, blue blocks are with our modified AdaFactor optimizer.
The shapes in the white area were not able to fit into memory in any setting.
For space reasons, we show here only the models pertaining to the experiments presented, but note that with our modifications we were able to fit thin ViT models of a depth up to 100 encoder blocks.

\begin{table}
  \setlength{\tabcolsep}{5pt}
  \setlength{\extrarowheight}{5pt}
  \renewcommand{\arraystretch}{0.75}
  \centering
  \caption{Model architecture details.}\label{tbl:models}
  \vspace{0.5em}
  \begin{tabulary}{1.0\linewidth}{LCCCCRRR}
    \toprule
    \multirow{3}{=}[3pt]{\centering \rotatebox{90}{\bf{Name}}} &
    \multirow{3}{=}[3pt]{\centering \rotatebox{90}{\bf{Width}}} &
    \multirow{3}{=}[3pt]{\centering \rotatebox{90}{\bf{Depth}}} &
    \multirow{3}{=}[3pt]{\centering \rotatebox{90}{\bf{MLP}}} &
    \multirow{3}{=}[3pt]{\centering \rotatebox{90}{\bf{Heads}}} &
    \multirow{3}{=}[3pt]{\centering \rotatebox{90}{\shortstack[c]{\bf{Mio.} \\ \bf{Param}}}} &
    \multicolumn{2}{c}{\bf{GFLOPs}} \\
    \cmidrule[0.5pt]{7-8}
     &  &  &  &  & & \multicolumn{1}{c}{$224^2$} & \multicolumn{1}{c}{$384^2$} \\
    \cmidrule[0.5pt]{1-8}
    s/28  & 256  & 6  & 1024 & 8  &  5.4 & 0.7   & 2.0 \\
    s/16  & 256  & 6  & 1024 & 8  &  5.0 & 2.2   & 7.8 \\
    S/32  & 384  & 12 & 1536 & 6  &   22 & 2.3   & 6.9 \\
    Ti/16 & 192  & 12 & 768  & 3  &  5.5 & 2.5   & 9.5 \\
    B/32  & 768  & 12 & 3072 & 12 &   87 & 8.7   & 26.0 \\
    S/16  & 384  & 12 & 1536 & 6  &   22 & 9.2   & 31.2 \\
    B/28  & 768  & 12 & 3072 & 12 &   87 & 11.3  & 30.5 \\
    B/16  & 768  & 12 & 3072 & 12 &   86 & 35.1  & 111.3 \\
    L/16  & 1024 & 24 & 4096 & 16 &  303 & 122.9 & 382.8 \\
    g/14  & 1408 & 40 & 6144 & 16 & 1011 & 533.1 & 1596.4 \\
    G/14  & 1664 & 48 & 8192 & 16 & 1843 & 965.3 & 2859.9 \\
    \bottomrule
  \end{tabulary}
\end{table}

The original Vision Transformer publication contains a study in Appendix~D2 about the trade-offs between scaling the different components, concluding that it is most effective to scale all aspects (depth, width, MLP-width, and patch-size) simultaneously and by a similar amount.
We follow this recommendation, and select shapes for ViT-g and ViT-G at the limit of what fits in memory accordingly, as shown in Figure~\ref{fig:shapefinder} and summarized in Table~\ref{tbl:models}.

\section{Related Work}\label{sec:relwork}

\textbf{Smaller Vision Transformers\,\,}
Early work on Transformers for vision focused on  small networks for CIFAR-10~\cite{cordonnier2020}.
The Vision Transformer~\cite{dosovitskiy2020}, however, was proposed in the context of state-of-the-art medium and large-scale image recognition; the smallest model (ViT-B) containing 86M parameters.
\cite{touvron2020training} present smaller ViT sizes for training from-scratch, down to ViT-Ti, with 5M parameters.
New variants of ViT introduce smaller and cheaper architectures.
For example, T2T-ViT~\cite{yuan2021tokens} reduces the number of parameters and compute using a new tokenization and narrower networks.
Pyramidal ViTs~\cite{wang2021pyramid}, designed for dense prediction tasks, follow a CNN-like pyramidal structure, that also reduces the size of the model.
Hybrids of CNNs and Transformers typically allow smaller models to perform well, such as the ViT-CNN hybrid in \cite{dosovitskiy2020}, BoTNet~\cite{srinivas2021bottleneck}, and HaloNet~\cite{vaswani2021scaling}.
However, the other direction, increasing the scale of ViT, is less explored.
While language Transformers are still much larger than Vision Transformers, understanding the scaling properties and the improvements introduced in this paper represent a step in this direction.

\textbf{Scaling Laws\,\,}
\cite{kaplan2020scaling} present a thorough study of the empirical scaling laws of neural language models.
The authors fit power laws that describe the relationships between compute, data size, model size, and performance. 
Following these laws, GPT-3, a 175B parameter language model was successfully trained~\cite{brown2020language}.
\cite{henighan2020scaling} presents laws for autoregressive generative modelling in other modalities, including the generation of images.
Our paper contains the first study of scaling laws for the discriminative modelling of images.

\textbf{Scaling-up Vision Models\,\,}
Many papers scale up CNNs to attain improved performance.
EfficientNets~\cite{efficientnet,efficientnet_v2} present a scaling strategy that balances compute between depth, width, and resolution and apply it to MobileNets.
This strategy is revisited in \cite{bello2021revisiting,resnet_strikes} to further improve the performance of ResNets~\cite{he2016deep}.
Large CNNs have attained excellent performance in visual recognition, such as AmoebaNet-B(18, 512) (557M parameters) trained using GPipe pipeline parallelism~\cite{gpipe}, ResNeXt-101 32×48d (829M parameters) pre-trained on weakly-labelled Instagram images~\cite{instagram}, EfficientNet-L2 (480M parameters) trained with ImageNet pseudo-labels on JFT-300M~\cite{noisystudent}, and BiT-L-ResNet152x4 (928M parameters) pre-trained on JFT-300M~\cite{kolesnikov2019big}.
Recently, ~\cite{deepvit,cait} explore strategies to scale the depth of ViTs. 
We are the first to scale Vision Transformers to even larger size and reache new state-of-the-art results doing so.
The concurrent work ~\cite{coatnet} focuses on CNN and ViT hybrid architectures.

\section{Discussion}\label{sec:discussion}

\textbf{Limitations.} This work uses the proprietary JFT-3B dataset for the scaling laws study. To make our insights more reliable and generalizable, we verify that the scaling laws also apply on the public ImageNet-21k dataset. 

\textbf{Societal  impact.} A potential broader cost of this work is the energy required to perform the experiments in our scaling study, especially in training the largest ViT-G model.
However, this cost may be amortized in two ways.
First, such studies of scaling laws need only be performed once; We hope future developers of ViT models may use our results to design models that can be trained with fewer compute resources.
Second, the models trained are designed primarily for transfer learning.
Transfer of pre-trained weights is much less expensive than training from scratch on a downstream task, and typically reaches higher accuracy.
Therefore, by transferring our models to many tasks, the pre-training compute is further amortized.

\section{Conclusion}\label{sec:conclusion}

We demonstrate that the performance-compute frontier for ViT models with enough training data roughly follows a (saturating) power law. 
Crucially, in order to stay on this frontier one has to simultaneously scale compute and model size; that is, not increasing a model's size when extra compute becomes available is suboptimal. 
We also demonstrate that larger models are much more sample efficient and are great few-shot learners. 
Finally, we present a new training recipe, which allows one to efficiently train large and high-performing ViT models. Note, that our conclusions may not necessarily generalize beyond the scale we have studied and they may not generalize beyond the ViT family of models.
\paragraph{Acknowledgements} We thank James Bradbury and Vivek Sharma for their help on using large-scale infrastructure; Alexey Dosovitskiy, Joan Puigcerver, Basil Mustafa, Carlos Riquelme for insightful discussions; Tom Duerig, Austin Tarango, Daniel Keysers, Howard Zhou, Wenlei Zhou, Yanan Bao for discussions on JFT; the Google Brain team at large for providing a supportive research environment.

{\small
\bibliographystyle{ieee_fullname}
\bibliography{scaling}

\begin{thebibliography}{10}\itemsep=-1pt

\bibitem{aka2021measuring}
Osman Aka, Ken Burke, Alex B{\"a}uerle, Christina Greer, and Margaret Mitchell.
\newblock Measuring model biases in the absence of ground truth.
\newblock {\em arXiv preprint arXiv:2103.03417}, 2021.

\bibitem{Barbu2019ObjectNetAL}
Andrei Barbu, D. Mayo, Julian Alverio, William Luo, Christopher Wang, Dan
  Gutfreund, J. Tenenbaum, and Boris Katz.
\newblock Objectnet: A large-scale bias-controlled dataset for pushing the
  limits of object recognition models.
\newblock In {\em NeurIPS}, 2019.

\bibitem{bello2021revisiting}
Irwan Bello, William Fedus, Xianzhi Du, Ekin~D Cubuk, Aravind Srinivas,
  Tsung-Yi Lin, Jonathon Shlens, and Barret Zoph.
\newblock Revisiting resnets: Improved training and scaling strategies.
\newblock {\em arXiv preprint arXiv:2103.07579}, 2021.

\bibitem{beyer2020imagenet}
Lucas Beyer, Olivier~J. Hénaff, Alexander Kolesnikov, Xiaohua Zhai, and Aäron
  van~den Oord.
\newblock Are we done with imagenet?
\newblock {\em arXiv preprint arXiv:2006.07159}, 2020.

\bibitem{big_vision}
Lucas Beyer, Xiaohua Zhai, and Alexander Kolesnikov.
\newblock Big vision.
\newblock \url{https://github.com/google-research/big_vision}, 2022.

\bibitem{gpt3}
Tom Brown, Benjamin Mann, Nick Ryder, Melanie Subbiah, Jared~D Kaplan, Prafulla
  Dhariwal, Arvind Neelakantan, Pranav Shyam, Girish Sastry, Amanda Askell,
  Sandhini Agarwal, Ariel Herbert-Voss, Gretchen Krueger, Tom Henighan, Rewon
  Child, Aditya Ramesh, Daniel Ziegler, Jeffrey Wu, Clemens Winter, Chris
  Hesse, Mark Chen, Eric Sigler, Mateusz Litwin, Scott Gray, Benjamin Chess,
  Jack Clark, Christopher Berner, Sam McCandlish, Alec Radford, Ilya Sutskever,
  and Dario Amodei.
\newblock Language models are few-shot learners.
\newblock In {\em NeurIPS}, 2020.

\bibitem{brown2020language}
Tom~B Brown, Benjamin Mann, Nick Ryder, Melanie Subbiah, Jared Kaplan, Prafulla
  Dhariwal, Arvind Neelakantan, Pranav Shyam, Girish Sastry, Amanda Askell,
  et~al.
\newblock Language models are few-shot learners.
\newblock {\em arXiv preprint arXiv:2005.14165}, 2020.

\bibitem{carion2020endtoend}
Nicolas Carion, Francisco Massa, Gabriel Synnaeve, Nicolas Usunier, Alexander
  Kirillov, and Sergey Zagoruyko.
\newblock End-to-end object detection with transformers.
\newblock {\em arXiv preprint arXiv:2005.12872}, 2020.

\bibitem{dino}
Mathilde Caron, Hugo Touvron, Ishan Misra, Herv{\'{e}} J{\'{e}}gou, Julien
  Mairal, Piotr Bojanowski, and Armand Joulin.
\newblock Emerging properties in self-supervised vision transformers.
\newblock {\em CoRR}, abs/2104.14294, 2021.

\bibitem{chen2020big}
Ting Chen, Simon Kornblith, Kevin Swersky, Mohammad Norouzi, and Geoffrey
  Hinton.
\newblock Big self-supervised models are strong semi-supervised learners.
\newblock {\em arXiv preprint arXiv:2006.10029}, 2020.

\bibitem{cordonnier2020}
Jean-Baptiste Cordonnier, Andreas Loukas, and Martin Jaggi.
\newblock On the relationship between self-attention and convolutional layers.
\newblock In {\em ICLR}, 2020.

\bibitem{cortes1995support}
Corinna Cortes and Vladimir Vapnik.
\newblock Support-vector networks.
\newblock {\em Machine learning}, 1995.

\bibitem{coatnet}
Zihang Dai, Hanxiao Liu, Quoc~V. Le, and Mingxing Tan.
\newblock Coatnet: Marrying convolution and attention for all data sizes.
\newblock {\em CoRR}, abs/2106.04803, 2021.

\bibitem{imagenet}
J. {Deng}, W. {Dong}, R. {Socher}, L. {Li}, {Kai Li}, and {Li Fei-Fei}.
\newblock Imagenet: A large-scale hierarchical image database.
\newblock In {\em CVPR}, 2009.

\bibitem{devlin2019bert}
Jacob Devlin, Ming-Wei Chang, Kenton Lee, and Kristina Toutanova.
\newblock Bert: Pre-training of deep bidirectional transformers for language
  understanding.
\newblock {\em arXiv preprint arXiv:1810.04805}, 2018.

\bibitem{dosovitskiy2020}
Alexey Dosovitskiy, Lucas Beyer, Alexander Kolesnikov, Dirk Weissenborn,
  Xiaohua Zhai, Thomas Unterthiner, Mostafa Dehghani, Matthias Minderer, Georg
  Heigold, Sylvain Gelly, Jakob Uszkoreit, and Neil Houlsby.
\newblock {An Image is Worth 16x16 Words: Transformers for Image Recognition at
  Scale}.
\newblock In {\em ICLR}, 2021.

\bibitem{grill2020bootstrap}
Jean-Bastien Grill, Florian Strub, Florent Altché, Corentin Tallec, Pierre~H.
  Richemond, Elena Buchatskaya, Carl Doersch, Bernardo~Avila Pires,
  Zhaohan~Daniel Guo, Mohammad~Gheshlaghi Azar, Bilal Piot, Koray Kavukcuoglu,
  Rémi Munos, and Michal Valko.
\newblock Bootstrap your own latent: A new approach to self-supervised
  learning.
\newblock {\em arXiv preprint arXiv:2006.07733}, 2020.

\bibitem{he2016deep}
Kaiming He, Xiangyu Zhang, Shaoqing Ren, and Jian Sun.
\newblock Deep residual learning for image recognition.
\newblock In {\em CVPR}, 2016.

\bibitem{henighan2020scaling}
Tom Henighan, Jared Kaplan, Mor Katz, Mark Chen, Christopher Hesse, Jacob
  Jackson, Heewoo Jun, Tom~B Brown, Prafulla Dhariwal, Scott Gray, et~al.
\newblock Scaling laws for autoregressive generative modeling.
\newblock {\em arXiv preprint arXiv:2010.14701}, 2020.

\bibitem{gpipe}
Yanping Huang, Youlong Cheng, Ankur Bapna, Orhan Firat, Dehao Chen, Mia Chen,
  HyoukJoong Lee, Jiquan Ngiam, Quoc~V Le, Yonghui Wu, and zhifeng Chen.
\newblock Gpipe: Efficient training of giant neural networks using pipeline
  parallelism.
\newblock In {\em NeurIPS}, 2019.

\bibitem{jia2021scaling}
Chao Jia, Yinfei Yang, Ye Xia, Yi-Ting Chen, Zarana Parekh, Hieu Pham, Quoc~V.
  Le, Yunhsuan Sung, Zhen Li, and Tom Duerig.
\newblock Scaling up visual and vision-language representation learning with
  noisy text supervision.
\newblock {\em arXiv preprint arXiv:2102.05918}, 2021.

\bibitem{kaplan2020scaling}
Jared Kaplan, Sam McCandlish, Tom Henighan, Tom~B Brown, Benjamin Chess, Rewon
  Child, Scott Gray, Alec Radford, Jeffrey Wu, and Dario Amodei.
\newblock Scaling laws for neural language models.
\newblock {\em arXiv preprint arXiv:2001.08361}, 2020.

\bibitem{kolesnikov2019big}
Alexander Kolesnikov, Lucas Beyer, Xiaohua Zhai, J. Puigcerver, Jessica Yung,
  S. Gelly, and N. Houlsby.
\newblock {Big Transfer (BiT): General Visual Representation Learning}.
\newblock In {\em ECCV}, 2020.

\bibitem{cifar}
Alex Krizhevsky.
\newblock Learning multiple layers of features from tiny images.
\newblock Technical report, 2009.

\bibitem{lee2019set}
Juho Lee, Yoonho Lee, Jungtaek Kim, Adam Kosiorek, Seungjin Choi, and Yee~Whye
  Teh.
\newblock Set transformer: A framework for attention-based
  permutation-invariant neural networks.
\newblock In {\em ICML}, 2019.

\bibitem{gshard}
Dmitry Lepikhin, HyoukJoong Lee, Yuanzhong Xu, Dehao Chen, Orhan Firat, Yanping
  Huang, Maxim Krikun, Noam Shazeer, and Zhifeng Chen.
\newblock Gshard: Scaling giant models with conditional computation and
  automatic sharding.
\newblock {\em arXiv preprint arXiv:2006.16668}, 2020.

\bibitem{instagram}
Dhruv Mahajan, Ross Girshick, Vignesh Ramanathan, Kaiming He, Manohar Paluri,
  Yixuan Li, Ashwin Bharambe, and Laurens van~der Maaten.
\newblock Exploring the limits of weakly supervised pretraining.
\newblock In {\em ECCV}, September 2018.

\bibitem{pets}
Omkar~M. Parkhi, Andrea Vedaldi, Andrew Zisserman, and C.~V. Jawahar.
\newblock Cats and dogs.
\newblock In {\em CVPR}, 2012.

\bibitem{pham2020meta}
Hieu Pham, Zihang Dai, Qizhe Xie, Minh-Thang Luong, and Quoc~V. Le.
\newblock Meta pseudo labels.
\newblock {\em arXiv preprint arXiv:2003.10580}, 2020.

\bibitem{polyak}
B.~T. Polyak and A.~B. Juditsky.
\newblock Acceleration of stochastic approximation by averaging.
\newblock {\em SIAM Journal on Control and Optimization}, 30(4):838--855, 1992.

\bibitem{radford2021learning}
Alec Radford, Jong~Wook Kim, Chris Hallacy, Aditya Ramesh, Gabriel Goh,
  Sandhini Agarwal, Girish Sastry, Amanda Askell, Pamela Mishkin, Jack Clark,
  Gretchen Krueger, and Ilya Sutskever.
\newblock Learning transferable visual models from natural language
  supervision.
\newblock {\em arXiv preprint arXiv:2103.00020}, 2021.

\bibitem{zero_optimizer}
Samyam Rajbhandari, Jeff Rasley, Olatunji Ruwase, and Yuxiong He.
\newblock Zero: memory optimizations toward training trillion parameter models.
\newblock In Christine Cuicchi, Irene Qualters, and William~T. Kramer, editors,
  {\em Proceedings of the International Conference for High Performance
  Computing, Networking, Storage and Analysis, {SC} 2020, Virtual Event /
  Atlanta, Georgia, USA, November 9-19, 2020}, page~20. {IEEE/ACM}, 2020.

\bibitem{recht2019imagenet}
Benjamin Recht, Rebecca Roelofs, Ludwig Schmidt, and Vaishaal Shankar.
\newblock Do imagenet classifiers generalize to imagenet?
\newblock {\em arXiv preprint arXiv:1902.10811}, 2019.

\bibitem{ILSVRC15}
Olga Russakovsky, Jia Deng, Hao Su, Jonathan Krause, Sanjeev Satheesh, Sean Ma,
  Zhiheng Huang, Andrej Karpathy, Aditya Khosla, Michael Bernstein,
  Alexander~C. Berg, and Li Fei-Fei.
\newblock {ImageNet Large Scale Visual Recognition Challenge}.
\newblock {\em IJCV}, 115(3):211--252, 2015.

\bibitem{adafactor}
Noam Shazeer and Mitchell Stern.
\newblock Adafactor: Adaptive learning rates with sublinear memory cost.
\newblock In {\em ICML}, 2018.

\bibitem{srinivas2021bottleneck}
Aravind Srinivas, Tsung-Yi Lin, Niki Parmar, Jonathon Shlens, Pieter Abbeel,
  and Ashish Vaswani.
\newblock Bottleneck transformers for visual recognition.
\newblock {\em arXiv preprint arXiv:2101.11605}, 2021.

\bibitem{sun2017unreasonable}
Chen Sun, Abhinav Shrivastava, Saurabh Singh, and Abhinav Gupta.
\newblock {Revisiting Unreasonable Effectiveness of Data in Deep Learning Era}.
\newblock {\em ICCV}, Oct 2017.

\bibitem{efficientnet}
Mingxing Tan and Quoc Le.
\newblock {E}fficient{N}et: Rethinking model scaling for convolutional neural
  networks.
\newblock In {\em ICML}, 2019.

\bibitem{efficientnet_v2}
Mingxing Tan and Quoc~V. Le.
\newblock Efficientnetv2: Smaller models and faster training.
\newblock In Marina Meila and Tong Zhang, editors, {\em Proceedings of the 38th
  International Conference on Machine Learning, {ICML} 2021, 18-24 July 2021,
  Virtual Event}, volume 139 of {\em Proceedings of Machine Learning Research},
  pages 10096--10106. {PMLR}, 2021.

\bibitem{adam_1bit}
Hanlin Tang, Shaoduo Gan, Ammar~Ahmad Awan, Samyam Rajbhandari, Conglong Li,
  Xiangru Lian, Ji Liu, Ce Zhang, and Yuxiong He.
\newblock 1-bit adam: Communication efficient large-scale training with adam's
  convergence speed.
\newblock In Marina Meila and Tong Zhang, editors, {\em Proceedings of the 38th
  International Conference on Machine Learning, {ICML} 2021, 18-24 July 2021,
  Virtual Event}, volume 139 of {\em Proceedings of Machine Learning Research},
  pages 10118--10129. {PMLR}, 2021.

\bibitem{touvron2020training}
Hugo Touvron, Matthieu Cord, Matthijs Douze, Francisco Massa, Alexandre
  Sablayrolles, and Herv{\'e} J{\'e}gou.
\newblock Training data-efficient image transformers \& distillation through
  attention.
\newblock {\em arXiv preprint arXiv:2012.12877}, 2020.

\bibitem{cait}
Hugo Touvron, Matthieu Cord, Alexandre Sablayrolles, Gabriel Synnaeve, and
  Herv{\'{e}} J{\'{e}}gou.
\newblock Going deeper with image transformers.
\newblock {\em CoRR}, abs/2103.17239, 2021.

\bibitem{touvron2020fixing}
Hugo Touvron, Andrea Vedaldi, Matthijs Douze, and Hervé Jégou.
\newblock Fixing the train-test resolution discrepancy.
\newblock {\em arXiv preprint arXiv:1906.06423}, 2020.

\bibitem{vaswani2021scaling}
Ashish Vaswani, Prajit Ramachandran, Aravind Srinivas, Niki Parmar, Blake
  Hechtman, and Jonathon Shlens.
\newblock Scaling local self-attention for parameter efficient visual
  backbones.
\newblock {\em arXiv preprint arXiv:2103.12731}, 2021.

\bibitem{vaswani2017attention}
Ashish Vaswani, Noam Shazeer, Niki Parmar, Jakob Uszkoreit, Llion Jones,
  Aidan~N. Gomez, Lukasz Kaiser, and Illia Polosukhin.
\newblock Attention is all you need.
\newblock {\em arXiv preprint arXiv:1706.03762}, 2017.

\bibitem{wang2021pyramid}
Wenhai Wang, Enze Xie, Xiang Li, Deng-Ping Fan, Kaitao Song, Ding Liang, Tong
  Lu, Ping Luo, and Ling Shao.
\newblock Pyramid vision transformer: A versatile backbone for dense prediction
  without convolutions.
\newblock {\em arXiv preprint arXiv:2102.12122}, 2021.

\bibitem{cub}
P. Welinder, S. Branson, T. Mita, C. Wah, F. Schroff, S. Belongie, and P.
  Perona.
\newblock {Caltech-UCSD Birds 200}.
\newblock Technical Report CNS-TR-2010-001, California Institute of Technology,
  2010.

\bibitem{resnet_strikes}
Ross Wightman, Hugo Touvron, and Herv{\'{e}} J{\'{e}}gou.
\newblock Resnet strikes back: An improved training procedure in timm.
\newblock {\em CoRR}, abs/2110.00476, 2021.

\bibitem{xie2019selftraining}
Qizhe Xie, Minh-Thang Luong, Eduard Hovy, and Quoc~V. Le.
\newblock Self-training with noisy student improves imagenet classification.
\newblock {\em arXiv preprint arXiv:1911.04252}, 2019.

\bibitem{noisystudent}
Qizhe Xie, Minh-Thang Luong, Eduard Hovy, and Quoc~V. Le.
\newblock Self-training with noisy student improves imagenet classification.
\newblock In {\em CVPR}, June 2020.

\bibitem{yuan2021tokens}
Li Yuan, Yunpeng Chen, Tao Wang, Weihao Yu, Yujun Shi, Zihang Jiang, Francis~EH
  Tay, Jiashi Feng, and Shuicheng Yan.
\newblock Tokens-to-token vit: Training vision transformers from scratch on
  imagenet.
\newblock {\em arXiv preprint arXiv:2101.11986}, 2021.

\bibitem{zhai2019s4l}
Xiaohua Zhai, Avital Oliver, Alexander Kolesnikov, and Lucas Beyer.
\newblock S4l: Self-supervised semi-supervised learning.
\newblock In {\em ICCV}, pages 1476--1485, 2019.

\bibitem{zhai2019largescale}
Xiaohua Zhai, Joan Puigcerver, Alexander Kolesnikov, Pierre Ruyssen, Carlos
  Riquelme, Mario Lucic, Josip Djolonga, Andre~Susano Pinto, Maxim Neumann,
  Alexey Dosovitskiy, Lucas Beyer, Olivier Bachem, Michael Tschannen, Marcin
  Michalski, Olivier Bousquet, Sylvain Gelly, and Neil Houlsby.
\newblock A large-scale study of representation learning with the visual task
  adaptation benchmark.
\newblock {\em arXiv preprint arXiv:1910.04867}, 2019.

\bibitem{deepvit}
Daquan Zhou, Bingyi Kang, Xiaojie Jin, Linjie Yang, Xiaochen Lian, Qibin Hou,
  and Jiashi Feng.
\newblock Deepvit: Towards deeper vision transformer.
\newblock {\em CoRR}, abs/2103.11886, 2021.

\end{thebibliography}
}

\clearpage{}
\appendix

\section{More few-shot transfer results}\label{sec:app:more_results}

\begin{figure*}[h]
  \begin{center}
    \includegraphics[width=\linewidth]{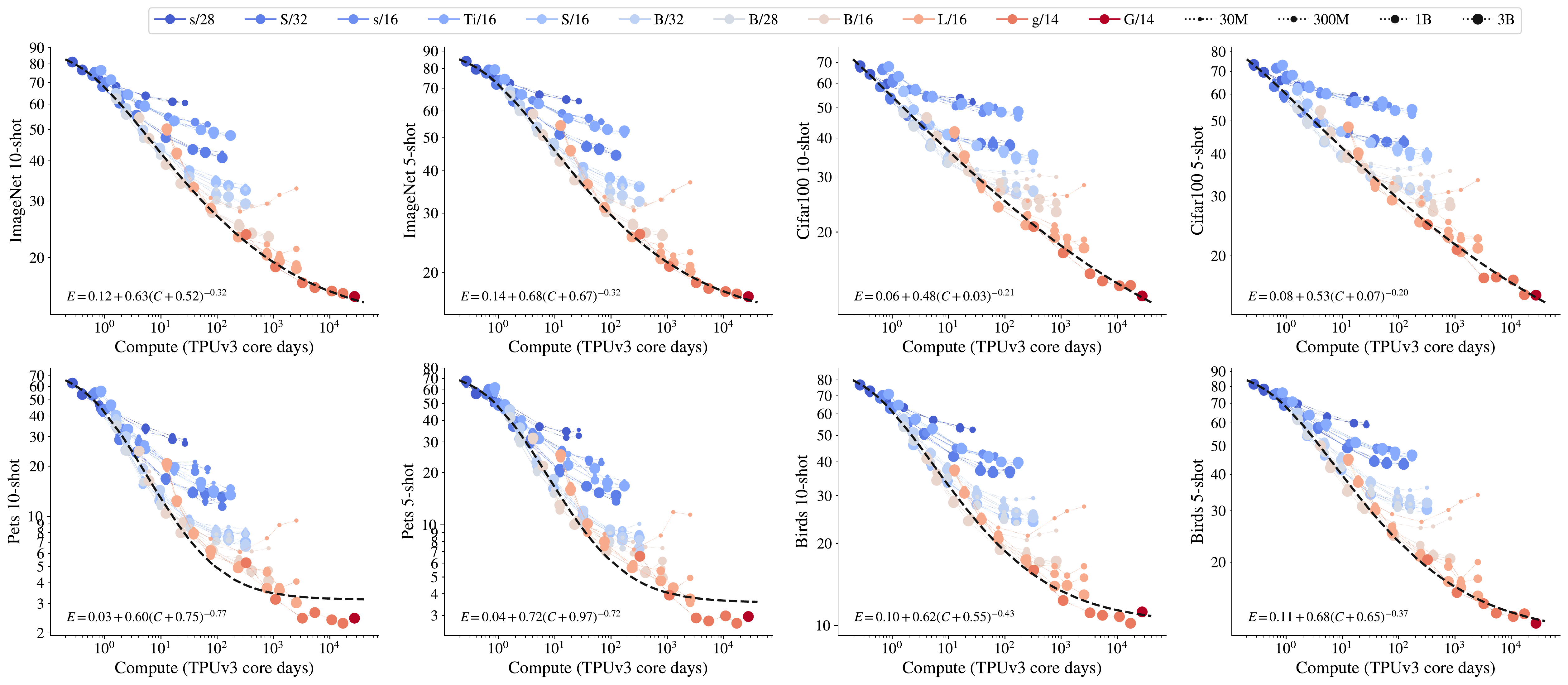}
  \end{center}
  \caption{Representation quality as a function of total training compute. Representation quality is measured as few-shot error rate on four datasets. 
  Sometimes, like in pets 5/10shot, the law does not fit the evidence perfectly, maybe the models are not ideal or the law is not universal.
  }
  \label{fig:scaling_laws_all_results}
\end{figure*}

We observe similar scaling laws on more datasets, including Oxford IIIT Pets~\cite{pets}, CIFAR-100~\cite{cifar}, and Caltech-UCSD Birds~\cite{cub}. The results are presented in Figure~\ref{fig:scaling_laws_all_results}.

\section{Pre-training details}\label{sec:app:adaptation_details}
We pre-train all the ViT models using adafactor optimizer with half-precision momentum. 
We use the default $\beta_1=0.9$ and $\beta_2=0.999$ (clipping threshold) for adafactor.
We use batch size 4096 for all the models smaller than \emph{ViT-g}.
For models \emph{ViT-g} and \emph{ViT-G}, to speed up the training, we scale up the batch size at most to 32\,768 and distribute the training to 2048 TPUv3 chips.
We set weight decay to 3.0 for the ``head'' and 0.03 for the ``body''.
All the models are pre-trained at resolution $224\times224$, with inception crop followed by random horizontal flip pre-process.
We use reciprocal square-root schedule with a linear learning rate warmup of 10k steps.
We cooldown the training at multiple steps as noted in Tables from Section~\ref{sec:app:table}.

\section{Configuration file for pre-training ViT-g}\label{app:configs}

We present the full configuration for training the ViT-g/14 model. It follows the \texttt{big\_vision} codebase\cite{big_vision} conventions.

\begin{lstlisting}[language=Python, caption=Full config for ViT-g/14 pre-training.]

def get_config():
  config = mlc.ConfigDict()

  config.dataset = 'jft_3b'
  config.val_split = 'val'
  config.train_split = 'train'
  config.num_classes = 29_593
  config.init_head_bias = -10.0

  # Fits 32 images per TPUv3 core with ViT-g/14.
  config.batch_size = 4096*4

  pp_common = '|value_range(-1, 1)'
  pp_common += f'|onehot({config.num_classes})'
  pp_common += '|keep("image", "labels")'
  config.pp_train = 'inception_crop(224)|flip_lr' + pp_common
  config.pp_eval = 'resize_small(256)|central_crop(224)' + pp_common
  config.shuffle_buffer_size = 250_000

  config.log_training_steps = 50
  config.log_eval_steps = 1000

  config.checkpoint_steps = 1000
  config.keep_checkpoint_steps = 10_000

  config.prefetch_to_device = 1
  config.trial = 0

  # Model section
  config.model_name = 'vit'
  config.model = mlc.ConfigDict()
  config.model.variant = 'g/14'
  config.model.pool_type = 'map'

  # Optimizer section
  config.optax_name = 'big_vision.scale_by_adafactor'
  config.grad_clip_norm = 1.0
  config.lr = 8e-4
  config.wd = 0.03 * 8e-4
  config.wd_mults = [
      ('.*head/kernel', 100.0),
      ('.*/kernel', 1.0),
  ]
  config.schedule = dict(
      decay_type='rsqrt', timescale=10_000, warmup_steps=10_000,
      cooldown_steps=50_000)
  config.total_steps = 1_000_000

  # Few-shot eval section
  config.fewshot = get_fewshot()
  config.fewshot.log_steps = 10_000

  return config

\end{lstlisting}

\section{Adaptation details}\label{sec:app:adaptation_details}
We report both the few-shot linear regression and finetune results on mutiple datasets.
For few-shot linear regression, we simply solve the $l2-$regularized linear regression problem, using the frozen embeddings extracted from $224\times224$ resolution images.

For finetune evaluation, we use SGD optimizer with momentum.
We use batch size 512 and gradient clipping at global norm 1. 
We do not use weight decay for finetuning. 
Following~\cite{kolesnikov2019big, touvron2020fixing}, we use higher resolution for finetuning. 
More specifically, we use $384\times384$ resolution for ViT models smaller than \emph{ViT-g}, and $518\times518$ resolution for both \emph{ViT-g} and \emph{ViT-G}.
We use Polyak averaging~\cite{polyak} only for the \emph{ViT-G} model during fine-tuning, similar to~\cite{dosovitskiy2020}.
We use a cosine learning rate schedule for 20k steps by default, except a flat learning rate for \emph{ViT-G} with Polyak averaging.
We linearly warm up the learning rate for 500 steps.
We sweep over two learning rates \{0.03, 0.01\} and choose the better one using a held-out 2\% training split.
On VTAB tasks, we use a fixed 0.01 learning rate with a cosine learning rate schedule. We train for 2\,500 steps in total. 

\section{Impact of resolution and patch size}\label{sec:app:resolution_and_patch_size}

In this section, we answer the question of ``what happens if we scale up the resolution, while keeping number of tokens fixed?".
We perform experiments on ImageNet-21K to verify this point, by scaling the resolution and patch size linearly together. 
We observed in Table~\ref{tbl:res_and_patch_size} that the quality difference is pretty subtle if we increase patch and resolution together. What matters for ViT architecture is the total number of patches, which has already been covered in Table~\ref{tbl:models} with different patch sizes: 32, 28, 16, and 14.

\vspace{-0.5em}
\begin{table}[h]
  \newcolumntype{C}{>{\centering\arraybackslash}X}
  \newcolumntype{R}{>{\raggedleft\arraybackslash}X}
  \setlength{\tabcolsep}{0pt}
  \setlength{\extrarowheight}{5pt}
  \renewcommand{\arraystretch}{0.75}
  \centering
  \caption{Results of different resolutions and patch sizes.}\label{tbl:res_and_patch_size}
  \vspace{-1em}
  \begin{tabularx}{\linewidth}{p{1.1cm}p{0.3cm}Cp{0.3cm}Cp{0.3cm}Cp{0.3cm}Cp{0.3cm}Cp{0.3cm}C}
    \toprule[1pt]
     \bf{Model} && B/32 && B/48 && B/64 && S/16 && S/24 && S/32 \\
     \bf{Res.} && 224 && 336 && 448 && 224 && 336 && 448 \\
    \midrule
     \bf{INet10} && 64.43 && 64.65 && 64.67 && 63.42 && 63.79 && 63.50 \\
    \bottomrule[1pt]
  \end{tabularx}
\end{table}
\vspace{-1em}

\section{Full table of few-shot results}\label{sec:app:table}

We provide the 5-shot learning and 10-shot learning results, on the four datasets from Figure~\ref{fig:scaling_laws_all_results}. 
Both \emph{ViT-g/14} and \emph{ViT-G/14} are summarized in Table~\ref{tbl:g_14}.
Note that we use at most 32\,768 batch size for \emph{ViT-g/14} and \emph{ViT-G/14}.
To make the following tables more readable, we normalize the number of steps in Table~\ref{tbl:g_14} assuming the batch size is always 4096, i.e. $Images\ Seen / 4096$.
All the other smaller ViT models are summarized in Table~\ref{tbl:l_16}, Table~\ref{tbl:b_16}, Table~\ref{tbl:b_28}, Table~\ref{tbl:b_32}, Table~\ref{tbl:s_16},  Table~\ref{tbl:ti_16}, Table~\ref{tbl:xs_16}, Table~\ref{tbl:s_32}, Table~\ref{tbl:xs_28}. 
We are aware of a few missing rows, which do not affect the trend for the scaling laws plot. 

\begin{table*}[h]
  \setlength{\tabcolsep}{5pt}
  \setlength{\extrarowheight}{5pt}
  \renewcommand{\arraystretch}{0.75}
  \centering
  \caption{Tabular representation of the few-shot results (\%) for model \emph{ViT-g/14} and \emph{ViT-G/14}.}\label{tbl:g_14}
  \begin{tabulary}{1.0\textwidth}{l r c c c c c c c c}
    \toprule[1pt]
    \bf{Data Size} & \bf{Steps} & \bf{INet5} & \bf{INet10} & \bf{Cifar5} & \bf{Cifar10} & \bf{Pets5} & \bf{Pets10} & \bf{Birds5} & \bf{Birds10} \\
    \midrule
\multicolumn{10}{c}{\emph{ViT-g/14}} \\
\arrayrulecolor{lightgray}\midrule[0.25pt]\arrayrulecolor{black}
3B    & 120K  & 74.0 & 76.4 & 75.3 & 79.2 & 93.4 & 94.7 & 79.6 & 84.0 \\
3B    & 400K  & 79.1 & 81.3 & 79.1 & 82.9 & 96.1 & 96.8 & 84.2 & 87.7 \\
3B    & 1.2M    & 81.3 & 83.3 & 82.8 & 85.3 & 97.1 & 97.5 & 85.6 & 88.9 \\
3B    & 2M    & 82.0 & 83.9 & 82.7 & 86.1 & 97.2 & 97.3 & 86.6 & 89.1 \\
3B    & 4M    & 82.4 & 84.3 & 83.0 & 86.5 & 97.0 & 97.6 & 87.0 & 89.2 \\
3B    & 6.3M    & 82.7 & 84.5 & 84.6 & 86.5 & 97.3 & 97.7 & 86.7 & 89.8 \\
    \midrule
\multicolumn{10}{c}{\emph{ViT-G/14}} \\
\arrayrulecolor{lightgray}\midrule[0.25pt]\arrayrulecolor{black}
3B    & 5M    & 83.0 & 84.9 & 84.7 & 87.5 & 97.0 & 97.5 & 87.6 & 88.8 \\
    \bottomrule
  \end{tabulary}
\end{table*}

\begin{table*}[h]
  \setlength{\tabcolsep}{5pt}
  \setlength{\extrarowheight}{5pt}
  \renewcommand{\arraystretch}{0.75}
  \centering
  \caption{Tabular representation of the few-shot results (\%) for model \emph{L/16}.}\label{tbl:l_16}
  \begin{tabulary}{1.0\textwidth}{l r c c c c c c c c}
    \toprule[1pt]
    \bf{Data Size} & \bf{Steps} & \bf{INet5} & \bf{INet10} & \bf{Cifar5} & \bf{Cifar10} & \bf{Pets5} & \bf{Pets10} & \bf{Birds5} & \bf{Birds10} \\
    \midrule
30M   & 20K   & 45.5 & 49.6 & 52.3 & 58.0 & 74.4 & 81.1 & 55.1 & 63.0 \\
30M   & 30K   & 53.3 & 57.4 & 58.8 & 64.5 & 82.8 & 87.0 & 61.1 & 68.3 \\
30M   & 60K   & 62.6 & 66.0 & 66.2 & 71.3 & 90.3 & 91.6 & 68.6 & 74.6 \\
30M   & 120K  & 66.5 & 69.3 & 67.2 & 72.8 & 91.2 & 92.7 & 70.8 & 77.1 \\
30M   & 400K  & 69.5 & 72.2 & 70.2 & 74.8 & 92.8 & 93.9 & 72.5 & 78.6 \\
30M   & 1.2M    & 67.1 & 70.6 & 69.2 & 73.6 & 91.3 & 92.6 & 69.8 & 75.2 \\
30M   & 2M    & 65.1 & 68.7 & 68.7 & 73.9 & 88.2 & 91.1 & 67.7 & 73.7 \\
30M   & 4M    & 63.1 & 67.2 & 66.6 & 71.7 & 88.6 & 90.6 & 66.0 & 72.6 \\
\arrayrulecolor{lightgray}\midrule[0.25pt]\arrayrulecolor{black}
300M  & 20K   & 45.0 & 49.7 & 52.1 & 57.3 & 74.4 & 79.1 & 55.5 & 63.2 \\
300M  & 30K   & 54.0 & 57.4 & 60.5 & 65.5 & 83.1 & 87.5 & 61.3 & 68.3 \\
300M  & 60K   & 63.0 & 66.5 & 68.0 & 72.8 & 90.2 & 92.2 & 68.7 & 74.7 \\
300M  & 120K  & 68.5 & 71.4 & 70.9 & 75.4 & 92.1 & 93.7 & 74.1 & 79.5 \\
300M  & 400K  & 74.6 & 77.0 & 71.5 & 77.1 & 94.2 & 95.1 & 78.8 & 83.3 \\
300M  & 1.2M    & 76.0 & 78.0 & 74.8 & 79.0 & 95.4 & 95.2 & 79.8 & 83.5 \\
300M  & 2M    & 77.5 & 79.5 & 77.0 & 81.5 & 95.8 & 96.3 & 82.5 & 84.8 \\
300M  & 4M    & 77.0 & 78.7 & 77.4 & 81.1 & 95.1 & 95.9 & 80.0 & 83.5 \\
\arrayrulecolor{lightgray}\midrule[0.25pt]\arrayrulecolor{black}
1B    & 20K   & 45.9 & 50.6 & 52.6 & 58.9 & 75.8 & 80.2 & 55.8 & 63.4 \\
1B    & 30K   & 54.7 & 58.4 & 60.8 & 66.0 & 84.5 & 88.0 & 62.4 & 69.5 \\
1B    & 60K   & 63.4 & 66.9 & 68.2 & 72.1 & 91.0 & 92.3 & 69.3 & 75.0 \\
1B    & 120K  & 68.5 & 71.3 & 70.7 & 75.7 & 92.5 & 94.2 & 73.5 & 78.4 \\
1B    & 400K  & 74.6 & 76.9 & 74.7 & 77.4 & 94.1 & 94.9 & 78.5 & 82.4 \\
1B    & 1.2M    & 77.1 & 79.2 & 76.3 & 79.8 & 94.6 & 95.4 & 80.7 & 84.5 \\
1B    & 2M    & 78.5 & 80.0 & 77.6 & 80.8 & 95.7 & 96.2 & 82.9 & 85.8 \\
1B    & 4M    & 79.1 & 81.0 & 77.5 & 82.3 & 96.5 & 96.9 & 82.4 & 85.0 \\
\arrayrulecolor{lightgray}\midrule[0.25pt]\arrayrulecolor{black}
3B    & 20K   & 45.7 & 49.7 & 51.9 & 58.1 & 74.6 & 79.2 & 54.9 & 62.6 \\
3B    & 30K   & 54.1 & 57.7 & 59.7 & 64.9 & 84.0 & 87.6 & 62.4 & 69.4 \\
3B    & 60K   & 63.6 & 66.9 & 67.2 & 71.5 & 89.8 & 92.1 & 69.9 & 75.5 \\
3B    & 120K  & 68.9 & 71.5 & 70.8 & 76.0 & 92.0 & 93.8 & 74.8 & 79.2 \\
3B    & 400K  & 74.5 & 76.8 & 74.8 & 78.9 & 94.4 & 95.1 & 79.0 & 82.6 \\
3B    & 1.2M    & 78.0 & 79.7 & 77.3 & 80.8 & 95.3 & 96.3 & 82.8 & 86.1 \\
3B    & 2M    & 78.6 & 80.5 & 79.4 & 82.4 & 95.7 & 96.4 & 83.6 & 86.0 \\
3B    & 4M    & 79.8 & 81.5 & 78.9 & 82.2 & 96.3 & 97.0 & 84.7 & 87.1 \\
    \bottomrule
  \end{tabulary}
\end{table*}

\begin{table*}[t]
  \setlength{\tabcolsep}{5pt}
  \setlength{\extrarowheight}{5pt}
  \renewcommand{\arraystretch}{0.75}
  \centering
  \caption{Tabular representation of the few-shot results (\%) for model \emph{B/16}.}\label{tbl:b_16}
  \begin{tabulary}{1.0\textwidth}{l r c c c c c c c c}
    \toprule[1pt]
    \bf{Data Size} & \bf{Steps} & \bf{INet5} & \bf{INet10} & \bf{Cifar5} & \bf{Cifar10} & \bf{Pets5} & \bf{Pets10} & \bf{Birds5} & \bf{Birds10} \\
    \midrule
30M   & 20K   & 40.8 & 45.4 & 47.2 & 53.5 & 68.6 & 74.8 & 49.3 & 57.8 \\
30M   & 30K   & 48.0 & 52.2 & 53.1 & 58.7 & 79.1 & 81.4 & 56.7 & 64.5 \\
30M   & 60K   & 56.6 & 60.5 & 60.1 & 64.8 & 87.0 & 88.6 & 64.1 & 71.3 \\
30M   & 120K  & 61.6 & 65.4 & 63.4 & 68.4 & 89.1 & 90.8 & 69.0 & 75.1 \\
30M   & 400K  & 67.1 & 70.4 & 66.0 & 70.8 & 91.5 & 93.3 & 72.8 & 78.2 \\
30M   & 1.2M    & 68.8 & 71.4 & 65.2 & 70.5 & 92.0 & 93.8 & 73.8 & 79.4 \\
30M   & 2M    & 68.2 & 71.1 & 65.7 & 69.7 & 92.5 & 93.8 & 73.5 & 78.4 \\
30M   & 4M    & 67.2 & 70.5 & 64.4 & 70.0 & 92.7 & 93.6 & 71.8 & 77.7 \\
\arrayrulecolor{lightgray}\midrule[0.25pt]\arrayrulecolor{black}
300M  & 20K   & 41.0 & 45.6 & 48.4 & 54.6 & 69.7 & 76.1 & 49.9 & 57.6 \\
300M  & 30K   & 48.6 & 52.6 & 53.5 & 58.7 & 78.5 & 83.1 & 56.5 & 64.2 \\
300M  & 60K   & 57.3 & 60.6 & 60.0 & 65.9 & 86.8 & 89.8 & 63.7 & 70.8 \\
300M  & 120K  & 62.3 & 65.5 & 63.8 & 69.4 & 89.3 & 91.1 & 68.9 & 75.3 \\
300M  & 400K  & 69.3 & 72.1 & 68.8 & 73.1 & 92.3 & 94.1 & 75.5 & 80.5 \\
300M  & 1.2M    & 72.3 & 74.9 & 68.4 & 72.2 & 93.1 & 94.4 & 77.7 & 81.1 \\
300M  & 2M    & 73.3 & 76.0 & 71.2 & 74.7 & 94.5 & 95.5 & 78.4 & 82.7 \\
300M  & 4M    & 73.6 & 76.3 & 70.8 & 74.9 & 94.4 & 95.4 & 79.2 & 82.7 \\
\arrayrulecolor{lightgray}\midrule[0.25pt]\arrayrulecolor{black}
1B    & 20K   & 40.9 & 45.0 & 47.0 & 53.0 & 69.3 & 75.8 & 49.4 & 58.1 \\
1B    & 30K   & 48.5 & 52.6 & 53.2 & 59.1 & 78.3 & 83.4 & 56.6 & 63.6 \\
1B    & 60K   & 57.3 & 61.0 & 60.0 & 65.2 & 87.9 & 89.6 & 65.0 & 71.0 \\
1B    & 120K  & 62.3 & 65.9 & 63.4 & 68.8 & 89.5 & 90.6 & 69.8 & 75.2 \\
1B    & 400K  & 69.2 & 72.2 & 67.9 & 71.1 & 93.0 & 93.6 & 74.6 & 80.5 \\
1B    & 1.2M    & 71.9 & 74.4 & 71.5 & 75.7 & 94.3 & 94.8 & 77.1 & 81.6 \\
1B    & 2M    & 73.8 & 75.9 & 72.2 & 76.2 & 94.8 & 95.2 & 79.0 & 82.8 \\
1B    & 4M    & 74.3 & 76.7 & 71.1 & 75.2 & 93.8 & 95.3 & 79.3 & 83.4 \\
\arrayrulecolor{lightgray}\midrule[0.25pt]\arrayrulecolor{black}
3B    & 20K   & 41.3 & 45.7 & 46.4 & 53.5 & 68.3 & 75.4 & 50.0 & 58.9 \\
3B    & 30K   & 49.2 & 53.1 & 54.0 & 59.5 & 78.2 & 83.6 & 57.9 & 65.8 \\
3B    & 60K   & 57.4 & 61.0 & 61.0 & 65.3 & 87.0 & 89.6 & 65.0 & 71.7 \\
3B    & 120K  & 62.5 & 66.0 & 63.8 & 68.6 & 90.0 & 92.0 & 68.9 & 75.8 \\
3B    & 400K  & 69.8 & 72.4 & 68.2 & 72.9 & 93.0 & 93.8 & 75.3 & 81.0 \\
3B    & 1.2M    & 72.3 & 74.9 & 71.3 & 75.3 & 94.3 & 94.7 & 78.9 & 83.2 \\
3B    & 2M    & 73.8 & 76.3 & 72.9 & 74.6 & 94.7 & 95.3 & 79.0 & 82.9 \\
3B    & 4M    & 74.3 & 76.8 & 71.9 & 76.8 & 95.1 & 95.9 & 79.4 & 82.8 \\
    \bottomrule
  \end{tabulary}
\end{table*}

\begin{table*}[t]
  \setlength{\tabcolsep}{5pt}
  \setlength{\extrarowheight}{5pt}
  \renewcommand{\arraystretch}{0.75}
  \centering
  \caption{Tabular representation of the few-shot results (\%) for model \emph{B/28}.}\label{tbl:b_28}
  \begin{tabulary}{1.0\textwidth}{l r c c c c c c c c}
    \toprule[1pt]
    \bf{Data Size} & \bf{Steps} & \bf{INet5} & \bf{INet10} & \bf{Cifar5} & \bf{Cifar10} & \bf{Pets5} & \bf{Pets10} & \bf{Birds5} & \bf{Birds10} \\
    \midrule
30M   & 20K   & 33.9 & 37.8 & 46.9 & 52.3 & 60.5 & 66.2 & 38.7 & 46.7 \\
30M   & 30K   & 40.1 & 44.3 & 51.3 & 57.1 & 70.7 & 74.2 & 45.1 & 51.9 \\
30M   & 60K   & 48.6 & 52.9 & 56.8 & 61.9 & 80.0 & 84.4 & 52.3 & 60.2 \\
30M   & 120K  & 54.2 & 58.3 & 59.6 & 66.0 & 84.1 & 87.2 & 56.7 & 64.6 \\
30M   & 400K  & 61.4 & 64.8 & 62.6 & 68.4 & 90.2 & 92.1 & 63.6 & 70.3 \\
30M   & 1.2M  & 64.2 & 67.4 & 63.9 & 69.3 & 91.0 & 92.2 & 66.4 & 73.9 \\
30M   & 2M    & 64.4 & 67.5 & 64.0 & 69.0 & 91.6 & 92.2 & 68.4 & 73.7 \\
\arrayrulecolor{lightgray}\midrule[0.25pt]\arrayrulecolor{black}
300M  & 20K   & 33.6 & 37.5 & 44.7 & 51.1 & 58.3 & 67.2 & 38.5 & 46.1 \\
300M  & 30K   & 40.0 & 44.6 & 51.7 & 57.0 & 70.4 & 75.1 & 44.2 & 52.0 \\
300M  & 60K   & 48.2 & 52.8 & 55.9 & 61.3 & 80.2 & 83.8 & 52.0 & 59.7 \\
300M  & 120K  & 54.4 & 58.3 & 60.9 & 65.9 & 84.2 & 88.6 & 57.6 & 65.5 \\
300M  & 400K  & 63.1 & 66.1 & 65.8 & 70.8 & 90.2 & 91.4 & 64.5 & 71.9 \\
300M  & 1.2M  & 66.5 & 69.6 & 68.2 & 72.1 & 92.3 & 92.9 & 68.4 & 74.7 \\
300M  & 2M    & 67.9 & 70.9 & 68.1 & 72.7 & 92.7 & 92.8 & 70.0 & 76.2 \\
\arrayrulecolor{lightgray}\midrule[0.25pt]\arrayrulecolor{black}
1B    & 20K   & 33.6 & 37.9 & 45.6 & 51.8 & 58.9 & 64.5 & 38.7 & 46.0 \\
1B    & 30K   & 39.8 & 44.6 & 50.9 & 56.6 & 70.5 & 75.5 & 45.0 & 51.4 \\
1B    & 60K   & 48.3 & 53.1 & 56.6 & 62.4 & 79.6 & 84.0 & 52.6 & 60.1 \\
1B    & 120K  & 54.8 & 58.5 & 61.2 & 66.9 & 84.9 & 88.1 & 57.9 & 65.0 \\
1B    & 400K  & 63.1 & 66.6 & 65.3 & 70.2 & 89.9 & 91.5 & 65.5 & 72.4 \\
1B    & 1.2M  & 67.1 & 69.8 & 66.3 & 70.8 & 92.0 & 92.9 & 69.2 & 75.6 \\
\arrayrulecolor{lightgray}\midrule[0.25pt]\arrayrulecolor{black}
3B    & 20K   & 33.3 & 37.6 & 45.6 & 51.9 & 58.3 & 65.1 & 38.5 & 46.3 \\
3B    & 30K   & 40.0 & 44.2 & 50.5 & 56.4 & 69.0 & 75.1 & 45.4 & 52.1 \\
    \bottomrule
  \end{tabulary}
\end{table*}

\begin{table*}[t]
  \setlength{\tabcolsep}{5pt}
  \setlength{\extrarowheight}{5pt}
  \renewcommand{\arraystretch}{0.75}
  \centering
  \caption{Tabular representation of the few-shot results (\%) for model \emph{B/32}.}\label{tbl:b_32}
  \begin{tabulary}{1.0\textwidth}{l r c c c c c c c c}
    \toprule[1pt]
    \bf{Data Size} & \bf{Steps} & \bf{INet5} & \bf{INet10} & \bf{Cifar5} & \bf{Cifar10} & \bf{Pets5} & \bf{Pets10} & \bf{Birds5} & \bf{Birds10} \\
    \midrule
30M   & 20K   & 30.6 & 34.8 & 44.2 & 50.2 & 54.7 & 62.6 & 34.7 & 42.0 \\
30M   & 30K   & 37.5 & 41.4 & 49.5 & 55.0 & 66.9 & 72.0 & 41.9 & 49.1 \\
30M   & 60K   & 46.1 & 50.0 & 55.9 & 60.7 & 77.0 & 81.0 & 48.8 & 57.1 \\
30M   & 120K  & 51.7 & 55.8 & 59.7 & 64.2 & 82.2 & 85.2 & 54.1 & 62.1 \\
30M   & 400K  & 59.5 & 63.4 & 63.3 & 68.3 & 88.5 & 90.6 & 62.3 & 68.3 \\
30M   & 1.2M    & 63.2 & 66.7 & 64.1 & 68.8 & 90.8 & 91.8 & 64.3 & 70.7 \\
30M   & 2M    & 63.8 & 66.7 & 63.6 & 68.8 & 90.6 & 91.3 & 64.7 & 70.9 \\
30M   & 4M    & 63.2 & 67.3 & 62.4 & 68.5 & 89.9 & 91.6 & 64.0 & 70.6 \\
\arrayrulecolor{lightgray}\midrule[0.25pt]\arrayrulecolor{black}
300M  & 20K   & 30.5 & 35.2 & 45.1 & 50.0 & 54.7 & 60.6 & 35.8 & 42.7 \\
300M  & 30K   & 37.3 & 41.0 & 50.2 & 55.6 & 64.4 & 70.2 & 40.8 & 48.0 \\
300M  & 60K   & 45.7 & 49.9 & 56.6 & 62.2 & 75.6 & 80.8 & 48.8 & 56.1 \\
300M  & 120K  & 51.9 & 55.8 & 61.2 & 66.0 & 82.3 & 86.2 & 53.5 & 61.2 \\
300M  & 400K  & 60.1 & 64.1 & 65.4 & 70.6 & 88.9 & 90.7 & 61.5 & 68.1 \\
300M  & 1.2M    & 64.0 & 67.7 & 66.3 & 71.4 & 90.9 & 91.9 & 65.1 & 71.2 \\
300M  & 2M    & 66.4 & 69.3 & 67.7 & 73.1 & 91.6 & 92.1 & 67.5 & 73.7 \\
300M  & 4M    & 67.5 & 70.1 & 68.4 & 73.1 & 91.7 & 92.2 & 68.3 & 74.1 \\
\arrayrulecolor{lightgray}\midrule[0.25pt]\arrayrulecolor{black}
1B    & 20K   & 30.6 & 35.2 & 43.9 & 50.4 & 57.1 & 61.9 & 35.1 & 41.9 \\
1B    & 30K   & 37.1 & 41.9 & 49.5 & 55.4 & 65.3 & 71.8 & 41.6 & 48.5 \\
1B    & 60K   & 46.3 & 50.2 & 56.0 & 61.3 & 76.2 & 80.3 & 48.3 & 56.7 \\
1B    & 120K  & 51.9 & 55.9 & 60.8 & 65.2 & 81.5 & 85.7 & 54.6 & 62.3 \\
1B    & 400K  & 60.8 & 64.5 & 65.7 & 70.3 & 88.6 & 90.7 & 62.5 & 69.1 \\
1B    & 1.2M    & 65.1 & 68.1 & 66.6 & 72.3 & 90.7 & 91.7 & 66.6 & 72.9 \\
1B    & 2M    & 66.1 & 69.4 & 68.1 & 71.8 & 91.4 & 92.7 & 66.9 & 74.1 \\
1B    & 4M    & 67.5 & 70.7 & 67.4 & 73.3 & 91.7 & 93.1 & 68.2 & 74.2 \\
\arrayrulecolor{lightgray}\midrule[0.25pt]\arrayrulecolor{black}
3B    & 20K   & 31.5 & 35.2 & 44.9 & 51.0 & 53.9 & 62.7 & 35.1 & 43.6 \\
3B    & 30K   & 37.6 & 41.8 & 50.1 & 55.9 & 63.7 & 70.0 & 42.0 & 48.8 \\
3B    & 60K   & 46.2 & 50.1 & 56.7 & 62.3 & 77.0 & 80.7 & 49.6 & 57.1 \\
3B    & 120K  & 52.0 & 56.4 & 60.5 & 66.6 & 81.7 & 85.6 & 55.2 & 62.5 \\
3B    & 400K  & 61.6 & 64.3 & 65.7 & 70.3 & 88.0 & 90.9 & 62.5 & 69.8 \\
3B    & 1.2M    & 65.3 & 68.7 & 67.7 & 72.6 & 90.6 & 92.0 & 67.3 & 73.0 \\
3B    & 2M    & 66.2 & 69.1 & 68.7 & 73.5 & 91.9 & 92.7 & 68.3 & 74.0 \\
3B    & 4M    & 67.6 & 70.6 & 70.0 & 72.9 & 92.7 & 93.4 & 68.1 & 75.1 \\
    \bottomrule
  \end{tabulary}
\end{table*}

\begin{table*}[t]
  \setlength{\tabcolsep}{5pt}
  \setlength{\extrarowheight}{5pt}
  \renewcommand{\arraystretch}{0.75}
  \centering
  \caption{Tabular representation of the few-shot results (\%) for model \emph{S/16}.}\label{tbl:s_16}
  \begin{tabulary}{1.0\textwidth}{l r c c c c c c c c}
    \toprule[1pt]
    \bf{Data Size} & \bf{Steps} & \bf{INet5} & \bf{INet10} & \bf{Cifar5} & \bf{Cifar10} & \bf{Pets5} & \bf{Pets10} & \bf{Birds5} & \bf{Birds10} \\
    \midrule
30M   & 20K   & 32.1 & 36.4 & 39.2 & 44.5 & 56.1 & 62.3 & 38.4 & 45.8 \\
30M   & 30K   & 39.1 & 43.3 & 46.3 & 51.4 & 67.2 & 74.1 & 46.8 & 53.7 \\
30M   & 60K   & 47.1 & 51.2 & 50.7 & 55.7 & 78.4 & 83.1 & 54.0 & 61.6 \\
30M   & 120K  & 51.8 & 55.4 & 53.8 & 59.1 & 82.2 & 86.1 & 58.8 & 66.3 \\
30M   & 400K  & 58.4 & 61.8 & 57.2 & 62.6 & 88.3 & 90.6 & 65.0 & 71.6 \\
30M   & 1.2M  & 61.1 & 64.4 & 57.1 & 63.2 & 90.5 & 91.3 & 68.4 & 73.7 \\
30M   & 2M    & 61.9 & 65.6 & 58.8 & 64.1 & 91.0 & 92.0 & 68.0 & 74.0 \\
30M   & 4M    & 63.2 & 66.5 & 58.7 & 65.2 & 91.3 & 92.1 & 69.7 & 74.7 \\
\arrayrulecolor{lightgray}\midrule[0.25pt]\arrayrulecolor{black}
300M  & 20K   & 31.3 & 36.4 & 38.9 & 43.9 & 57.4 & 62.2 & 37.4 & 45.4 \\
300M  & 60K   & 47.4 & 51.1 & 50.8 & 56.6 & 78.3 & 83.8 & 53.4 & 62.0 \\
300M  & 120K  & 52.6 & 56.4 & 53.0 & 58.1 & 83.1 & 87.0 & 58.7 & 66.2 \\
300M  & 400K  & 58.9 & 62.7 & 56.1 & 61.5 & 88.6 & 90.2 & 65.6 & 72.2 \\
300M  & 1.2M  & 62.3 & 66.0 & 58.2 & 63.9 & 90.9 & 91.8 & 69.5 & 74.9 \\
300M  & 2M    & 63.3 & 66.4 & 60.1 & 64.9 & 91.9 & 93.0 & 69.9 & 75.5 \\
300M  & 4M    & 64.5 & 67.5 & 61.1 & 65.9 & 92.1 & 93.3 & 70.3 & 75.4 \\
\arrayrulecolor{lightgray}\midrule[0.25pt]\arrayrulecolor{black}
1B    & 20K   & 31.8 & 35.8 & 37.1 & 42.9 & 56.1 & 63.3 & 38.2 & 46.0 \\
1B    & 30K   & 39.2 & 43.1 & 44.2 & 50.3 & 68.1 & 75.1 & 44.6 & 52.6 \\
1B    & 60K   & 47.2 & 51.3 & 50.6 & 55.8 & 78.2 & 84.5 & 53.9 & 60.8 \\
1B    & 120K  & 51.5 & 55.8 & 53.6 & 59.4 & 83.3 & 86.8 & 59.0 & 66.8 \\
1B    & 400K  & 58.9 & 62.6 & 56.7 & 62.3 & 88.0 & 90.6 & 65.7 & 72.2 \\
1B    & 1.2M  & 61.5 & 65.2 & 59.6 & 64.4 & 90.7 & 92.1 & 67.3 & 75.0 \\
1B    & 2M    & 62.8 & 66.6 & 60.8 & 66.0 & 90.7 & 92.0 & 69.1 & 75.5 \\
1B    & 4M    & 64.0 & 67.4 & 61.4 & 66.2 & 91.2 & 92.1 & 69.5 & 74.9 \\
\arrayrulecolor{lightgray}\midrule[0.25pt]\arrayrulecolor{black}
3B    & 20K   & 32.3 & 36.5 & 38.4 & 43.8 & 56.2 & 59.7 & 37.7 & 45.4 \\
3B    & 30K   & 38.8 & 43.1 & 43.8 & 50.7 & 68.8 & 75.1 & 45.9 & 53.9 \\
3B    & 120K  & 52.6 & 56.3 & 53.5 & 58.8 & 83.8 & 87.6 & 58.5 & 66.0 \\
3B    & 400K  & 59.1 & 62.7 & 56.9 & 62.2 & 88.7 & 90.8 & 65.8 & 72.1 \\
3B    & 1.2M  & 62.1 & 65.7 & 58.7 & 63.6 & 91.0 & 92.2 & 68.6 & 74.9 \\
3B    & 2M    & 63.7 & 66.5 & 59.5 & 65.3 & 91.4 & 92.5 & 68.8 & 75.7 \\
3B    & 4M    & 64.1 & 67.6 & 60.3 & 64.6 & 91.6 & 93.0 & 69.7 & 75.9 \\
    \bottomrule
  \end{tabulary}
\end{table*}

\begin{table*}[t]
  \setlength{\tabcolsep}{5pt}
  \setlength{\extrarowheight}{5pt}
  \renewcommand{\arraystretch}{0.75}
  \centering
  \caption{Tabular representation of the few-shot results (\%) for model \emph{Ti/16}.}\label{tbl:ti_16}
  \begin{tabulary}{1.0\textwidth}{l r c c c c c c c c}
    \toprule[1pt]
    \bf{Data Size} & \bf{Steps} & \bf{INet5} & \bf{INet10} & \bf{Cifar5} & \bf{Cifar10} & \bf{Pets5} & \bf{Pets10} & \bf{Birds5} & \bf{Birds10} \\
    \midrule
30M   & 20K   & 20.2 & 23.5 & 26.9 & 32.0 & 37.8 & 44.6 & 24.2 & 29.2 \\
30M   & 30K   & 25.7 & 28.6 & 31.9 & 38.0 & 50.1 & 54.9 & 29.7 & 35.5 \\
30M   & 60K   & 32.9 & 36.0 & 36.6 & 41.7 & 61.9 & 65.7 & 36.9 & 43.1 \\
30M   & 120K  & 36.5 & 40.6 & 39.1 & 44.3 & 67.2 & 74.2 & 41.1 & 48.0 \\
30M   & 400K  & 42.3 & 46.7 & 41.7 & 48.2 & 76.1 & 82.0 & 47.4 & 55.2 \\
30M   & 1.2M    & 45.6 & 49.8 & 45.4 & 50.6 & 80.8 & 84.9 & 52.3 & 58.9 \\
30M   & 2M    & 47.4 & 50.5 & 45.1 & 51.0 & 81.0 & 84.1 & 53.6 & 59.4 \\
30M   & 4M    & 48.2 & 51.6 & 46.3 & 52.7 & 82.2 & 85.6 & 53.3 & 59.5 \\
\arrayrulecolor{lightgray}\midrule[0.25pt]\arrayrulecolor{black}
300M  & 20K   & 20.7 & 23.7 & 28.0 & 32.6 & 43.3 & 45.3 & 23.8 & 30.0 \\
300M  & 30K   & 25.8 & 28.8 & 32.2 & 36.8 & 49.7 & 55.0 & 29.6 & 35.6 \\
300M  & 60K   & 33.1 & 36.4 & 37.6 & 42.7 & 62.2 & 68.2 & 37.3 & 44.0 \\
300M  & 120K  & 37.3 & 41.1 & 39.9 & 45.4 & 67.0 & 75.1 & 42.6 & 48.2 \\
300M  & 400K  & 43.2 & 47.3 & 42.9 & 49.3 & 74.4 & 81.7 & 47.7 & 55.6 \\
300M  & 1.2M    & 46.4 & 50.8 & 45.6 & 51.7 & 81.7 & 85.4 & 51.3 & 59.1 \\
300M  & 2M    & 48.0 & 51.6 & 46.1 & 51.8 & 82.3 & 85.7 & 53.3 & 60.2 \\
300M  & 4M    & 49.0 & 51.9 & 46.6 & 52.6 & 83.1 & 86.3 & 54.3 & 61.4 \\
\arrayrulecolor{lightgray}\midrule[0.25pt]\arrayrulecolor{black}
1B    & 20K   & 20.4 & 23.5 & 27.7 & 32.8 & 40.5 & 45.3 & 24.0 & 29.8 \\
1B    & 30K   & 26.0 & 28.6 & 31.7 & 37.5 & 54.3 & 54.9 & 29.5 & 35.4 \\
1B    & 60K   & 32.7 & 36.0 & 36.1 & 42.1 & 59.5 & 66.7 & 36.1 & 42.8 \\
1B    & 120K  & 36.4 & 40.2 & 39.2 & 45.0 & 68.2 & 73.1 & 41.3 & 47.8 \\
1B    & 400K  & 42.9 & 47.2 & 43.9 & 49.3 & 77.8 & 80.8 & 47.9 & 54.6 \\
1B    & 1.2M    & 46.4 & 49.9 & 44.9 & 50.2 & 81.9 & 85.1 & 52.1 & 59.1 \\
1B    & 2M    & 47.5 & 51.8 & 46.3 & 51.9 & 83.7 & 86.4 & 53.9 & 60.2 \\
1B    & 4M    & 48.3 & 52.2 & 47.8 & 53.4 & 83.5 & 85.3 & 54.3 & 60.2 \\
\arrayrulecolor{lightgray}\midrule[0.25pt]\arrayrulecolor{black}
3B    & 20K   & 20.6 & 23.6 & 26.9 & 32.2 & 38.2 & 43.6 & 24.0 & 29.0 \\
3B    & 30K   & 25.6 & 28.5 & 31.7 & 36.9 & 50.6 & 53.4 & 29.2 & 35.3 \\
3B    & 60K   & 32.9 & 35.7 & 37.3 & 43.1 & 63.1 & 66.7 & 36.2 & 42.5 \\
3B    & 120K  & 37.1 & 41.0 & 40.0 & 45.9 & 68.8 & 74.6 & 40.5 & 47.0 \\
3B    & 400K  & 42.9 & 46.7 & 42.7 & 48.3 & 78.0 & 80.3 & 48.7 & 55.4 \\
3B    & 1.2M    & 46.0 & 50.1 & 43.0 & 49.8 & 78.9 & 84.0 & 50.9 & 58.2 \\
3B    & 2M    & 47.1 & 50.8 & 46.3 & 51.6 & 82.5 & 85.7 & 52.3 & 60.0 \\
3B    & 4M    & 47.6 & 52.1 & 45.9 & 51.3 & 83.2 & 86.6 & 53.4 & 60.2 \\
    \bottomrule
  \end{tabulary}
\end{table*}

\begin{table*}[t]
  \setlength{\tabcolsep}{5pt}
  \setlength{\extrarowheight}{5pt}
  \renewcommand{\arraystretch}{0.75}
  \centering
  \caption{Tabular representation of the few-shot results (\%) for model \emph{s/16}.}\label{tbl:xs_16}
  \begin{tabulary}{1.0\textwidth}{l r c c c c c c c c}
    \toprule[1pt]
    \bf{Data Size} & \bf{Steps} & \bf{INet5} & \bf{INet10} & \bf{Cifar5} & \bf{Cifar10} & \bf{Pets5} & \bf{Pets10} & \bf{Birds5} & \bf{Birds10} \\
    \midrule
30M   & 20K   & 20.3 & 23.8 & 27.2 & 32.7 & 36.8 & 47.1 & 24.2 & 30.0 \\
30M   & 30K   & 26.3 & 29.6 & 32.3 & 37.9 & 47.6 & 54.4 & 30.2 & 37.4 \\
30M   & 60K   & 32.7 & 36.0 & 36.9 & 43.6 & 58.9 & 65.3 & 36.6 & 44.0 \\
30M   & 120K  & 36.2 & 39.7 & 39.0 & 45.7 & 65.0 & 69.9 & 42.2 & 48.7 \\
30M   & 400K  & 40.9 & 44.9 & 41.9 & 48.1 & 74.2 & 79.0 & 48.7 & 55.3 \\
30M   & 1.2M  & 43.2 & 47.6 & 43.7 & 49.5 & 75.9 & 78.7 & 51.3 & 57.3 \\
30M   & 2M    & 43.7 & 48.2 & 43.8 & 49.2 & 76.3 & 81.6 & 51.2 & 59.0 \\
\arrayrulecolor{lightgray}\midrule[0.25pt]\arrayrulecolor{black}
300M  & 20K   & 20.4 & 23.7 & 28.1 & 32.7 & 42.3 & 43.9 & 24.6 & 29.7 \\
300M  & 30K   & 25.5 & 29.3 & 33.5 & 39.2 & 49.6 & 55.0 & 30.2 & 36.1 \\
300M  & 60K   & 31.4 & 35.3 & 37.5 & 43.4 & 59.3 & 66.7 & 37.0 & 43.8 \\
300M  & 120K  & 35.8 & 39.4 & 38.4 & 44.6 & 65.3 & 71.2 & 41.6 & 48.9 \\
300M  & 400K  & 40.7 & 44.7 & 42.5 & 48.8 & 72.6 & 79.2 & 48.6 & 55.6 \\
300M  & 1.2M  & 43.5 & 47.3 & 43.8 & 50.1 & 77.2 & 80.0 & 51.8 & 57.8 \\
300M  & 2M    & 44.3 & 48.1 & 44.4 & 50.2 & 77.3 & 80.6 & 51.2 & 57.8 \\
\arrayrulecolor{lightgray}\midrule[0.25pt]\arrayrulecolor{black}
1B    & 20K   & 20.6 & 24.0 & 27.8 & 32.8 & 38.5 & 44.1 & 23.7 & 30.1 \\
1B    & 30K   & 26.1 & 29.9 & 31.7 & 37.7 & 49.3 & 55.9 & 30.3 & 37.3 \\
1B    & 60K   & 32.1 & 36.3 & 36.8 & 42.6 & 60.3 & 66.1 & 37.0 & 43.6 \\
1B    & 120K  & 35.5 & 40.1 & 40.0 & 46.1 & 66.0 & 72.2 & 41.6 & 48.9 \\
1B    & 400K  & 41.0 & 45.2 & 43.0 & 49.4 & 73.3 & 79.1 & 48.8 & 55.0 \\
1B    & 1.2M  & 42.8 & 47.2 & 45.0 & 51.8 & 76.3 & 81.5 & 50.7 & 57.2 \\
\arrayrulecolor{lightgray}\midrule[0.25pt]\arrayrulecolor{black}
3B    & 20K   & 20.7 & 24.5 & 28.4 & 33.8 & 39.7 & 44.9 & 24.8 & 29.8 \\
3B    & 30K   & 26.2 & 30.0 & 33.5 & 39.6 & 51.2 & 56.2 & 29.5 & 36.4 \\
    \bottomrule
  \end{tabulary}
\end{table*}

\begin{table*}[t]
  \setlength{\tabcolsep}{5pt}
  \setlength{\extrarowheight}{5pt}
  \renewcommand{\arraystretch}{0.75}
  \centering
  \caption{Tabular representation of the few-shot results (\%) for model \emph{S/32}.}\label{tbl:s_32}
  \begin{tabulary}{1.0\textwidth}{l r c c c c c c c c}
    \toprule[1pt]
    \bf{Data Size} & \bf{Steps} & \bf{INet5} & \bf{INet10} & \bf{Cifar5} & \bf{Cifar10} & \bf{Pets5} & \bf{Pets10} & \bf{Birds5} & \bf{Birds10} \\
    \midrule
30M   & 20K   & 23.0 & 26.7 & 34.1 & 40.3 & 43.3 & 48.1 & 26.1 & 32.2 \\
30M   & 30K   & 28.8 & 32.6 & 40.0 & 46.3 & 51.9 & 59.6 & 31.2 & 37.1 \\
30M   & 60K   & 36.4 & 39.9 & 45.9 & 51.8 & 64.5 & 71.5 & 39.0 & 45.2 \\
30M   & 120K  & 40.3 & 44.4 & 49.8 & 55.7 & 71.0 & 76.9 & 43.7 & 50.7 \\
30M   & 400K  & 47.9 & 51.7 & 54.8 & 60.1 & 79.5 & 83.1 & 50.5 & 57.6 \\
30M   & 1.2M  & 51.6 & 55.9 & 55.2 & 60.7 & 84.0 & 86.6 & 54.3 & 60.6 \\
30M   & 2M    & 52.5 & 56.6 & 56.0 & 60.1 & 84.9 & 88.1 & 55.7 & 62.5 \\
30M   & 4M    & 54.5 & 57.6 & 55.3 & 61.3 & 86.0 & 87.9 & 56.7 & 63.2 \\
\arrayrulecolor{lightgray}\midrule[0.25pt]\arrayrulecolor{black}
300M  & 20K   & 23.1 & 27.0 & 36.1 & 41.5 & 42.6 & 47.0 & 25.8 & 32.0 \\
300M  & 30K   & 28.4 & 31.9 & 42.2 & 47.1 & 48.9 & 58.9 & 30.3 & 36.6 \\
300M  & 60K   & 35.0 & 39.3 & 47.7 & 52.5 & 62.9 & 69.5 & 36.9 & 44.8 \\
300M  & 120K  & 40.8 & 44.9 & 50.4 & 55.4 & 71.4 & 75.6 & 43.4 & 50.3 \\
300M  & 400K  & 48.4 & 52.4 & 54.5 & 59.9 & 79.3 & 83.9 & 50.6 & 57.5 \\
300M  & 1.2M  & 52.9 & 56.2 & 57.3 & 62.6 & 83.0 & 85.5 & 54.6 & 61.8 \\
300M  & 2M    & 53.4 & 57.4 & 57.1 & 62.9 & 84.5 & 87.7 & 55.2 & 62.4 \\
300M  & 4M    & 55.2 & 58.5 & 57.6 & 62.8 & 85.4 & 87.1 & 55.5 & 62.6 \\
\arrayrulecolor{lightgray}\midrule[0.25pt]\arrayrulecolor{black}
1B    & 30K   & 28.3 & 32.2 & 41.4 & 47.1 & 50.2 & 56.2 & 29.9 & 36.6 \\
1B    & 60K   & 35.7 & 39.7 & 47.1 & 53.1 & 63.4 & 70.0 & 36.9 & 44.7 \\
1B    & 120K  & 40.8 & 44.7 & 50.7 & 56.0 & 68.6 & 75.3 & 43.2 & 50.1 \\
1B    & 400K  & 48.3 & 52.4 & 54.0 & 59.6 & 80.2 & 83.4 & 50.5 & 57.7 \\
1B    & 1.2M  & 52.6 & 56.7 & 55.8 & 61.1 & 83.2 & 86.4 & 55.7 & 62.4 \\
1B    & 2M    & 54.3 & 58.0 & 56.7 & 61.2 & 84.9 & 86.6 & 56.3 & 63.7 \\
1B    & 4M    & 55.4 & 58.8 & 56.3 & 61.6 & 86.4 & 88.6 & 56.5 & 64.0 \\
\arrayrulecolor{lightgray}\midrule[0.25pt]\arrayrulecolor{black}
3B    & 20K   & 22.5 & 26.3 & 36.8 & 41.7 & 43.4 & 46.8 & 25.1 & 31.5 \\
3B    & 30K   & 28.2 & 32.0 & 40.4 & 46.2 & 49.6 & 56.9 & 31.2 & 37.2 \\
3B    & 60K   & 36.0 & 39.6 & 46.5 & 52.0 & 63.1 & 71.2 & 37.5 & 44.8 \\
3B    & 120K  & 40.5 & 44.5 & 50.6 & 56.0 & 70.0 & 75.9 & 42.1 & 49.1 \\
3B    & 400K  & 48.8 & 52.8 & 53.5 & 59.6 & 79.1 & 83.2 & 50.8 & 58.2 \\
3B    & 1.2M  & 52.9 & 56.6 & 55.9 & 61.6 & 83.3 & 86.1 & 55.7 & 62.9 \\
3B    & 2M    & 53.6 & 57.5 & 56.6 & 62.0 & 84.7 & 86.2 & 56.4 & 63.5 \\
3B    & 4M    & 55.6 & 59.1 & 56.8 & 62.2 & 85.2 & 86.9 & 56.8 & 63.2 \\
    \bottomrule
  \end{tabulary}
\end{table*}

\begin{table*}[t]
  \setlength{\tabcolsep}{5pt}
  \setlength{\extrarowheight}{5pt}
  \renewcommand{\arraystretch}{0.75}
  \centering
  \caption{Tabular representation of the few-shot results (\%) for model \emph{s/28}.}\label{tbl:xs_28}
  \begin{tabulary}{1.0\textwidth}{l r c c c c c c c c}
    \toprule[1pt]
    \bf{Data Size} & \bf{Steps} & \bf{INet5} & \bf{INet10} & \bf{Cifar5} & \bf{Cifar10} & \bf{Pets5} & \bf{Pets10} & \bf{Birds5} & \bf{Birds10} \\
    \midrule
30M   & 20K   & 16.0 & 18.9 & 24.9 & 31.9 & 37.0 & 36.5 & 18.6 & 24.6 \\
30M   & 30K   & 20.3 & 23.4 & 30.5 & 35.9 & 40.4 & 46.8 & 23.0 & 28.2 \\
30M   & 60K   & 24.6 & 28.4 & 34.7 & 41.1 & 48.7 & 54.4 & 28.2 & 34.3 \\
30M   & 120K  & 27.7 & 32.0 & 37.4 & 43.3 & 51.6 & 58.5 & 29.7 & 37.2 \\
30M   & 400K  & 32.0 & 36.3 & 39.1 & 45.2 & 62.0 & 68.1 & 35.7 & 42.9 \\
30M   & 1.2M  & 34.8 & 38.6 & 40.8 & 46.3 & 66.5 & 70.1 & 40.1 & 46.1 \\
30M   & 2M    & 35.9 & 39.3 & 41.9 & 47.0 & 64.7 & 71.3 & 39.7 & 47.4 \\
\arrayrulecolor{lightgray}\midrule[0.25pt]\arrayrulecolor{black}
300M  & 20K   & 16.5 & 19.1 & 26.8 & 31.9 & 32.9 & 35.8 & 19.7 & 23.8 \\
300M  & 30K   & 19.9 & 23.2 & 29.9 & 36.2 & 42.0 & 44.0 & 23.7 & 29.1 \\
300M  & 60K   & 24.8 & 28.4 & 34.9 & 41.3 & 50.3 & 56.0 & 28.5 & 33.6 \\
300M  & 120K  & 27.6 & 31.6 & 37.0 & 43.2 & 54.5 & 58.4 & 32.0 & 37.5 \\
300M  & 400K  & 32.9 & 36.6 & 39.4 & 45.4 & 63.9 & 65.5 & 37.3 & 43.1 \\
300M  & 1.2M  & 35.4 & 39.3 & 41.2 & 47.7 & 68.2 & 69.8 & 40.3 & 45.9 \\
300M  & 2M    & 35.9 & 39.5 & 41.8 & 47.9 & 67.4 & 72.7 & 41.1 & 47.5 \\
\arrayrulecolor{lightgray}\midrule[0.25pt]\arrayrulecolor{black}
1B    & 20K   & 16.0 & 19.0 & 27.6 & 33.1 & 34.3 & 37.9 & 19.1 & 24.1 \\
1B    & 30K   & 20.2 & 23.3 & 30.1 & 35.9 & 41.3 & 45.8 & 23.2 & 27.3 \\
1B    & 60K   & 24.5 & 28.2 & 33.9 & 39.9 & 47.1 & 53.3 & 26.6 & 32.8 \\
1B    & 120K  & 27.6 & 31.8 & 36.5 & 43.3 & 53.6 & 60.3 & 30.2 & 36.6 \\
1B    & 400K  & 33.0 & 36.3 & 39.8 & 45.5 & 63.1 & 66.4 & 37.1 & 43.0 \\
1B    & 1.2M  & 35.2 & 38.9 & 40.8 & 46.3 & 65.4 & 71.2 & 40.2 & 46.9 \\
\arrayrulecolor{lightgray}\midrule[0.25pt]\arrayrulecolor{black}
3B    & 20K   & 16.0 & 18.9 & 26.6 & 31.8 & 32.5 & 37.1 & 18.6 & 23.2 \\
3B    & 30K   & 20.4 & 23.3 & 30.4 & 36.0 & 43.0 & 46.2 & 21.7 & 27.0 \\
    \bottomrule
  \end{tabulary}
\end{table*}

\section{Full table of finetune results}\label{sec:app:table}
We provide the finetune results on ImageNet, as well as the results evaluated on the other two ImageNet V2 and ImageNet ReaL test splits. Results for all the models could be found from Table~\ref{tbl:l_16_ft}, Table~\ref{tbl:b_16_ft}, Table~\ref{tbl:b_28_ft}, Table~\ref{tbl:b_32_ft}, Table~\ref{tbl:s_16_ft}, Table~\ref{tbl:ti_16_ft}, Table~\ref{tbl:xs_16_ft}, Table~\ref{tbl:s_32_ft}, Table~\ref{tbl:xs_28_ft}. 
We are aware of a few missing rows, which do not affect the trend for the scaling laws plot. 
We show the total steps and the cooldown steps for each model, as well as the best finetune learning rate selected on ImageNet held-out 2\% training split.

\begin{table*}[h]
  \setlength{\tabcolsep}{5pt}
  \setlength{\extrarowheight}{5pt}
  \renewcommand{\arraystretch}{0.75}
  \centering
  \caption{Tabular representation of the finetune results (\%) for model \emph{ViT-L/16} on ImageNet, ImageNet V2 test set and ImageNet ReaL test set.}\label{tbl:l_16_ft}
  \begin{tabulary}{1.0\textwidth}{l r c c c c c c c c}
    \toprule[1pt]
    \bf{Data Size} & \bf{Steps} & \bf{Cooldown} & \bf{LR} & \bf{ImageNet} & \bf{ImageNet V2} & \bf{ImageNet ReaL} \\
    \midrule
30M   & 20K   & 10K   & 0.03 & 75.4 & 63.3 & 82.1 \\
30M   & 30K   & 10K   & 0.03 & 78.8 & 67.5 & 85.0 \\
30M   & 60K   & 10K   & 0.03 & 82.4 & 72.5 & 87.6 \\
30M   & 120K  & 50K   & 0.03 & 83.8 & 74.8 & 88.3 \\
30M   & 400K  & 50K   & 0.03 & 85.5 & 76.5 & 89.0 \\
30M   & 1.2M  & 50K   & 0.03 & 85.3 & 76.0 & 88.7 \\
30M   & 2M    & 50K   & 0.03 & 85.1 & 76.2 & 88.7 \\
30M   & 4M    & 50K   & 0.01 & 85.6 & 77.0 & 89.1 \\
\arrayrulecolor{lightgray}\midrule[0.25pt]\arrayrulecolor{black}
300M  & 20K   & 10K   & 0.03 & 75.1 & 63.5 & 81.9 \\
300M  & 30K   & 10K   & 0.03 & 79.1 & 67.7 & 85.2 \\
300M  & 60K   & 10K   & 0.03 & 82.7 & 72.9 & 87.9 \\
300M  & 120K  & 50K   & 0.03 & 84.7 & 75.4 & 89.1 \\
300M  & 400K  & 50K   & 0.03 & 86.5 & 77.5 & 89.8 \\
300M  & 1.2M  & 50K   & 0.03 & 87.3 & 78.8 & 89.8 \\
300M  & 2M    & 50K   & 0.03 & 87.7 & 78.6 & 89.8 \\
300M  & 4M    & 50K   & 0.01 & 88.0 & 79.5 & 90.3 \\
\arrayrulecolor{lightgray}\midrule[0.25pt]\arrayrulecolor{black}
1B    & 20K   & 10K   & 0.03 & 75.9 & 63.9 & 82.7 \\
1B    & 30K   & 10K   & 0.03 & 79.5 & 68.4 & 85.5 \\
1B    & 60K   & 10K   & 0.03 & 82.5 & 72.6 & 87.8 \\
1B    & 120K  & 50K   & 0.03 & 84.5 & 75.4 & 88.9 \\
1B    & 400K  & 50K   & 0.03 & 86.7 & 78.3 & 89.8 \\
1B    & 1.2M  & 50K   & 0.03 & 87.2 & 78.6 & 89.8 \\
1B    & 2M    & 50K   & 0.03 & 87.9 & 78.9 & 90.0 \\
1B    & 4M    & 50K   & 0.03 & 88.0 & 79.5 & 90.1 \\
\arrayrulecolor{lightgray}\midrule[0.25pt]\arrayrulecolor{black}
3B    & 20K   & 10K   & 0.03 & 75.5 & 63.0 & 82.2 \\
3B    & 30K   & 10K   & 0.03 & 79.3 & 68.3 & 85.4 \\
3B    & 60K   & 10K   & 0.03 & 82.7 & 73.5 & 87.7 \\
3B    & 120K  & 50K   & 0.03 & 84.7 & 75.6 & 89.0 \\
3B    & 400K  & 50K   & 0.03 & 87.0 & 78.5 & 90.1 \\
3B    & 1.2M  & 50K   & 0.03 & 87.8 & 79.4 & 90.0 \\
3B    & 2M    & 50K   & 0.03 & 87.9 & 79.6 & 90.0 \\
3B    & 4M    & 50K   & 0.01 & 88.5 & 80.4 & 90.4 \\
    \bottomrule
  \end{tabulary}
\end{table*}

\begin{table*}[h]
  \setlength{\tabcolsep}{5pt}
  \setlength{\extrarowheight}{5pt}
  \renewcommand{\arraystretch}{0.75}
  \centering
  \caption{Tabular representation of the finetune results (\%) for model \emph{ViT-B/16} on ImageNet, ImageNet V2 test set and ImageNet ReaL test set.}\label{tbl:b_16_ft}
  \begin{tabulary}{1.0\textwidth}{l r c c c c c c c c}
    \toprule[1pt]
    \bf{Data Size} & \bf{Steps} & \bf{Cooldown} & \bf{LR} & \bf{ImageNet} & \bf{ImageNet V2} & \bf{ImageNet ReaL} \\
    \midrule
30M   & 20K   & 10K   & 0.03 & 73.0 & 60.4 & 80.0 \\
30M   & 30K   & 10K   & 0.03 & 76.9 & 64.9 & 83.4 \\
30M   & 60K   & 10K   & 0.03 & 80.5 & 69.5 & 86.1 \\
30M   & 120K  & 50K   & 0.03 & 82.2 & 72.3 & 87.4 \\
30M   & 400K  & 50K   & 0.03 & 84.4 & 74.6 & 88.5 \\
30M   & 1.2M  & 50K   & 0.03 & 84.9 & 75.0 & 88.7 \\
30M   & 2M    & 50K   & 0.03 & 84.8 & 74.8 & 88.6 \\
30M   & 4M    & 50K   & 0.01 & 84.9 & 75.3 & 88.8 \\
\arrayrulecolor{lightgray}\midrule[0.25pt]\arrayrulecolor{black}
300M  & 20K   & 10K   & 0.03 & 73.5 & 61.0 & 80.5 \\
300M  & 30K   & 10K   & 0.03 & 77.2 & 65.2 & 83.8 \\
300M  & 60K   & 10K   & 0.03 & 80.6 & 69.9 & 86.3 \\
300M  & 120K  & 50K   & 0.03 & 82.3 & 72.5 & 87.5 \\
300M  & 400K  & 50K   & 0.03 & 84.9 & 75.5 & 89.0 \\
300M  & 1.2M  & 50K   & 0.03 & 86.0 & 76.7 & 89.4 \\
300M  & 2M    & 50K   & 0.01 & 86.2 & 76.8 & 89.5 \\
300M  & 4M    & 50K   & 0.01 & 86.7 & 77.6 & 89.7 \\
\arrayrulecolor{lightgray}\midrule[0.25pt]\arrayrulecolor{black}
1B    & 20K   & 10K   & 0.03 & 73.2 & 60.7 & 80.2 \\
1B    & 30K   & 10K   & 0.03 & 77.0 & 65.7 & 83.6 \\
1B    & 60K   & 10K   & 0.03 & 80.6 & 70.7 & 86.4 \\
1B    & 120K  & 50K   & 0.03 & 82.3 & 72.0 & 87.5 \\
1B    & 400K  & 50K   & 0.03 & 85.1 & 75.2 & 89.1 \\
1B    & 1.2M  & 50K   & 0.03 & 86.0 & 77.0 & 89.5 \\
1B    & 2M    & 50K   & 0.03 & 86.5 & 77.3 & 89.6 \\
1B    & 4M    & 50K   & 0.01 & 86.8 & 77.5 & 89.8 \\
\arrayrulecolor{lightgray}\midrule[0.25pt]\arrayrulecolor{black}
3B    & 20K   & 10K   & 0.03 & 73.4 & 61.0 & 80.4 \\
3B    & 30K   & 10K   & 0.03 & 77.1 & 65.5 & 83.7 \\
3B    & 60K   & 10K   & 0.03 & 80.5 & 70.0 & 86.2 \\
3B    & 120K  & 50K   & 0.03 & 82.5 & 72.7 & 87.6 \\
3B    & 400K  & 50K   & 0.03 & 85.1 & 75.7 & 89.1 \\
3B    & 1.2M  & 50K   & 0.03 & 86.0 & 77.1 & 89.4 \\
3B    & 2M    & 50K   & 0.03 & 86.3 & 77.0 & 89.6 \\
3B    & 4M    & 50K   & 0.03 & 86.6 & 77.4 & 89.7 \\
    \bottomrule
  \end{tabulary}
\end{table*}

\begin{table*}[h]
  \setlength{\tabcolsep}{5pt}
  \setlength{\extrarowheight}{5pt}
  \renewcommand{\arraystretch}{0.75}
  \centering
  \caption{Tabular representation of the finetune results (\%) for model \emph{ViT-B/28} on ImageNet, ImageNet V2 test set and ImageNet ReaL test set.}\label{tbl:b_28_ft}
  \begin{tabulary}{1.0\textwidth}{l r c c c c c c c c}
    \toprule[1pt]
    \bf{Data Size} & \bf{Steps} & \bf{Cooldown} & \bf{LR} & \bf{ImageNet} & \bf{ImageNet V2} & \bf{ImageNet ReaL} \\
    \midrule
30M   & 20K   & 10K   & 0.03 & 68.8 & 55.6 & 76.1 \\
30M   & 30K   & 10K   & 0.03 & 72.8 & 59.6 & 79.8 \\
30M   & 60K   & 10K   & 0.03 & 76.7 & 64.5 & 83.4 \\
30M   & 120K  & 50K   & 0.03 & 79.1 & 68.3 & 85.3 \\
30M   & 400K  & 50K   & 0.03 & 82.2 & 72.1 & 87.4 \\
30M   & 1.2M  & 50K   & 0.03 & 83.3 & 73.1 & 87.8 \\
30M   & 2M    & 50K   & 0.03 & 83.5 & 73.4 & 87.8 \\
\arrayrulecolor{lightgray}\midrule[0.25pt]\arrayrulecolor{black}
300M  & 20K   & 10K   & 0.03 & 68.9 & 56.0 & 76.2 \\
300M  & 30K   & 10K   & 0.03 & 72.8 & 60.2 & 80.0 \\
300M  & 60K   & 10K   & 0.03 & 77.0 & 65.0 & 83.5 \\
300M  & 120K  & 50K   & 0.03 & 79.4 & 68.2 & 85.3 \\
300M  & 400K  & 50K   & 0.03 & 82.8 & 72.6 & 87.7 \\
300M  & 1.2M  & 50K   & 0.03 & 84.1 & 74.6 & 88.5 \\
300M  & 2M    & 50K   & 0.03 & 84.4 & 74.6 & 88.5 \\
\arrayrulecolor{lightgray}\midrule[0.25pt]\arrayrulecolor{black}
1B    & 20K   & 10K   & 0.03 & 68.6 & 55.5 & 75.9 \\
1B    & 30K   & 10K   & 0.03 & 72.8 & 60.1 & 79.9 \\
1B    & 60K   & 10K   & 0.03 & 76.9 & 65.1 & 83.6 \\
1B    & 120K  & 50K   & 0.03 & 79.4 & 69.0 & 85.5 \\
1B    & 400K  & 50K   & 0.03 & 82.7 & 73.0 & 87.6 \\
1B    & 1.2M  & 50K   & 0.03 & 84.0 & 74.4 & 88.3 \\
\arrayrulecolor{lightgray}\midrule[0.25pt]\arrayrulecolor{black}
3B    & 20K   & 10K   & 0.03 & 68.8 & 55.3 & 75.9 \\
3B    & 30K   & 10K   & 0.03 & 72.6 & 60.2 & 79.7 \\
    \bottomrule
  \end{tabulary}
\end{table*}

\begin{table*}[h]
  \setlength{\tabcolsep}{5pt}
  \setlength{\extrarowheight}{5pt}
  \renewcommand{\arraystretch}{0.75}
  \centering
  \caption{Tabular representation of the finetune results (\%) for model \emph{ViT-B/32} on ImageNet, ImageNet V2 test set and ImageNet ReaL test set.}\label{tbl:b_32_ft}
  \begin{tabulary}{1.0\textwidth}{l r c c c c c c c c}
    \toprule[1pt]
    \bf{Data Size} & \bf{Steps} & \bf{Cooldown} & \bf{LR} & \bf{ImageNet} & \bf{ImageNet V2} & \bf{ImageNet ReaL} \\
    \midrule
30M   & 20K   & 10K   & 0.03 & 66.6 & 53.8 & 73.8 \\
30M   & 30K   & 10K   & 0.03 & 71.0 & 57.9 & 78.0 \\
30M   & 60K   & 10K   & 0.03 & 75.6 & 63.5 & 82.3 \\
30M   & 120K  & 50K   & 0.03 & 78.0 & 66.4 & 84.3 \\
30M   & 400K  & 50K   & 0.03 & 81.4 & 70.8 & 86.8 \\
30M   & 1.2M  & 50K   & 0.03 & 82.7 & 72.4 & 87.5 \\
30M   & 2M    & 50K   & 0.03 & 83.1 & 72.7 & 87.7 \\
30M   & 4M    & 50K   & 0.01 & 83.0 & 72.8 & 87.7 \\
\arrayrulecolor{lightgray}\midrule[0.25pt]\arrayrulecolor{black}
300M  & 20K   & 10K   & 0.03 & 66.6 & 53.4 & 73.9 \\
300M  & 30K   & 10K   & 0.03 & 70.8 & 58.0 & 78.0 \\
300M  & 60K   & 10K   & 0.03 & 75.5 & 63.2 & 82.2 \\
300M  & 120K  & 50K   & 0.03 & 78.3 & 66.7 & 84.5 \\
300M  & 400K  & 50K   & 0.03 & 81.8 & 71.4 & 87.0 \\
300M  & 1.2M  & 50K   & 0.03 & 83.3 & 73.4 & 87.9 \\
300M  & 2M    & 50K   & 0.03 & 83.7 & 73.9 & 88.2 \\
300M  & 4M    & 50K   & 0.01 & 83.9 & 74.3 & 88.3 \\
\arrayrulecolor{lightgray}\midrule[0.25pt]\arrayrulecolor{black}
1B    & 20K   & 10K   & 0.03 & 66.8 & 53.7 & 74.1 \\
1B    & 30K   & 10K   & 0.03 & 71.1 & 58.5 & 78.1 \\
1B    & 60K   & 10K   & 0.03 & 75.5 & 63.1 & 82.2 \\
1B    & 120K  & 50K   & 0.03 & 78.5 & 66.9 & 84.7 \\
1B    & 400K  & 50K   & 0.03 & 82.0 & 71.6 & 87.2 \\
1B    & 1.2M  & 50K   & 0.03 & 83.4 & 73.5 & 87.9 \\
1B    & 2M    & 50K   & 0.03 & 83.7 & 73.9 & 88.1 \\
1B    & 4M    & 50K   & 0.03 & 84.1 & 74.4 & 88.4 \\
\arrayrulecolor{lightgray}\midrule[0.25pt]\arrayrulecolor{black}
3B    & 20K   & 10K   & 0.03 & 66.7 & 53.7 & 73.9 \\
3B    & 30K   & 10K   & 0.03 & 71.0 & 58.4 & 78.1 \\
3B    & 60K   & 10K   & 0.03 & 75.6 & 63.4 & 82.3 \\
3B    & 120K  & 50K   & 0.03 & 78.3 & 67.3 & 84.6 \\
3B    & 400K  & 50K   & 0.03 & 82.2 & 71.8 & 87.1 \\
3B    & 1.2M  & 50K   & 0.03 & 83.5 & 73.5 & 87.9 \\
3B    & 2M    & 50K   & 0.03 & 83.8 & 74.0 & 88.2 \\
3B    & 4M    & 50K   & 0.03 & 84.1 & 74.4 & 88.2 \\
    \bottomrule
  \end{tabulary}
\end{table*}

\begin{table*}[h]
  \setlength{\tabcolsep}{5pt}
  \setlength{\extrarowheight}{5pt}
  \renewcommand{\arraystretch}{0.75}
  \centering
  \caption{Tabular representation of the finetune results (\%) for model \emph{ViT-S/16} on ImageNet, ImageNet V2 test set and ImageNet ReaL test set.}\label{tbl:s_16_ft}
  \begin{tabulary}{1.0\textwidth}{l r c c c c c c c c}
    \toprule[1pt]
    \bf{Data Size} & \bf{Steps} & \bf{Cooldown} & \bf{LR} & \bf{ImageNet} & \bf{ImageNet V2} & \bf{ImageNet ReaL} \\
    \midrule
30M   & 20K   & 10K   & 0.03 & 67.4 & 54.5 & 74.7 \\
30M   & 30K   & 10K   & 0.03 & 72.5 & 59.9 & 79.6 \\
30M   & 60K   & 10K   & 0.03 & 76.8 & 65.0 & 83.2 \\
30M   & 120K  & 50K   & 0.03 & 78.8 & 67.8 & 85.1 \\
30M   & 400K  & 50K   & 0.03 & 81.5 & 70.9 & 87.1 \\
30M   & 1.2M  & 50K   & 0.03 & 82.5 & 72.0 & 87.7 \\
30M   & 2M    & 50K   & 0.03 & 82.8 & 72.2 & 87.8 \\
30M   & 4M    & 50K   & 0.01 & 83.5 & 72.8 & 88.2 \\
\arrayrulecolor{lightgray}\midrule[0.25pt]\arrayrulecolor{black}
300M  & 20K   & 10K   & 0.03 & 67.8 & 54.8 & 75.0 \\
300M  & 30K   & 10K   & 0.03 & 72.6 & 60.3 & 79.7 \\
300M  & 60K   & 10K   & 0.03 & 76.8 & 65.3 & 83.4 \\
300M  & 120K  & 50K   & 0.03 & 79.0 & 68.0 & 85.3 \\
300M  & 400K  & 50K   & 0.03 & 81.7 & 71.2 & 87.3 \\
300M  & 1.2M  & 50K   & 0.03 & 82.9 & 72.9 & 87.9 \\
300M  & 2M    & 50K   & 0.01 & 83.3 & 73.4 & 88.3 \\
300M  & 4M    & 50K   & 0.01 & 83.9 & 74.2 & 88.5 \\
\arrayrulecolor{lightgray}\midrule[0.25pt]\arrayrulecolor{black}
1B    & 20K   & 10K   & 0.03 & 67.3 & 54.5 & 74.6 \\
1B    & 30K   & 10K   & 0.03 & 72.3 & 60.0 & 79.6 \\
1B    & 60K   & 10K   & 0.03 & 76.6 & 64.9 & 83.4 \\
1B    & 120K  & 50K   & 0.03 & 78.8 & 67.9 & 85.2 \\
1B    & 400K  & 50K   & 0.03 & 81.9 & 70.6 & 87.3 \\
1B    & 1.2M  & 50K   & 0.03 & 82.8 & 72.4 & 87.8 \\
1B    & 2M    & 50K   & 0.01 & 83.2 & 72.8 & 88.2 \\
1B    & 4M    & 50K   & 0.03 & 83.5 & 72.7 & 88.3 \\
\arrayrulecolor{lightgray}\midrule[0.25pt]\arrayrulecolor{black}
3B    & 20K   & 10K   & 0.03 & 67.5 & 54.5 & 74.9 \\
3B    & 30K   & 10K   & 0.03 & 72.3 & 60.0 & 79.4 \\
3B    & 60K   & 10K   & 0.03 & 76.7 & 64.8 & 83.3 \\
3B    & 120K  & 50K   & 0.03 & 79.1 & 67.9 & 85.4 \\
3B    & 400K  & 50K   & 0.03 & 81.9 & 71.1 & 87.3 \\
3B    & 1.2M  & 50K   & 0.01 & 82.8 & 72.7 & 87.9 \\
3B    & 2M    & 50K   & 0.03 & 83.2 & 73.2 & 88.1 \\
3B    & 4M    & 50K   & 0.01 & 84.0 & 73.8 & 88.5 \\
    \bottomrule
  \end{tabulary}
\end{table*}

\begin{table*}[h]
  \setlength{\tabcolsep}{5pt}
  \setlength{\extrarowheight}{5pt}
  \renewcommand{\arraystretch}{0.75}
  \centering
  \caption{Tabular representation of the finetune results (\%) for model \emph{ViT-Ti/16} on ImageNet, ImageNet V2 test set and ImageNet ReaL test set.}\label{tbl:ti_16_ft}
  \begin{tabulary}{1.0\textwidth}{l r c c c c c c c c}
    \toprule[1pt]
    \bf{Data Size} & \bf{Steps} & \bf{Cooldown} & \bf{LR} & \bf{ImageNet} & \bf{ImageNet V2} & \bf{ImageNet ReaL} \\
    \midrule
30M   & 20K   & 10K   & 0.03 & 55.5 & 43.6 & 62.5 \\
30M   & 30K   & 10K   & 0.03 & 61.8 & 49.2 & 69.2 \\
30M   & 60K   & 10K   & 0.03 & 67.8 & 55.2 & 75.5 \\
30M   & 120K  & 50K   & 0.03 & 71.2 & 58.6 & 78.5 \\
30M   & 400K  & 50K   & 0.03 & 74.9 & 62.8 & 82.1 \\
30M   & 1.2M  & 50K   & 0.01 & 76.5 & 64.5 & 83.4 \\
30M   & 2M    & 50K   & 0.03 & 76.7 & 64.7 & 83.4 \\
30M   & 4M    & 50K   & 0.01 & 77.5 & 65.6 & 84.2 \\
\arrayrulecolor{lightgray}\midrule[0.25pt]\arrayrulecolor{black}
300M  & 20K   & 10K   & 0.03 & 55.9 & 43.7 & 62.9 \\
300M  & 30K   & 10K   & 0.03 & 61.7 & 49.3 & 69.0 \\
300M  & 60K   & 10K   & 0.03 & 68.5 & 55.7 & 76.0 \\
300M  & 120K  & 50K   & 0.03 & 71.4 & 58.8 & 78.7 \\
300M  & 400K  & 50K   & 0.03 & 75.2 & 62.8 & 82.2 \\
300M  & 1.2M  & 50K   & 0.03 & 76.7 & 64.7 & 83.7 \\
300M  & 2M    & 50K   & 0.01 & 77.1 & 65.5 & 84.1 \\
300M  & 4M    & 50K   & 0.01 & 77.8 & 66.1 & 84.4 \\
\arrayrulecolor{lightgray}\midrule[0.25pt]\arrayrulecolor{black}
1B    & 20K   & 10K   & 0.03 & 55.8 & 43.2 & 62.8 \\
1B    & 30K   & 10K   & 0.03 & 61.6 & 49.1 & 69.0 \\
1B    & 60K   & 10K   & 0.03 & 67.9 & 54.8 & 75.4 \\
1B    & 120K  & 50K   & 0.03 & 71.1 & 58.3 & 78.5 \\
1B    & 400K  & 50K   & 0.03 & 74.9 & 63.0 & 82.1 \\
1B    & 1.2M  & 50K   & 0.03 & 76.7 & 64.6 & 83.6 \\
1B    & 2M    & 50K   & 0.01 & 77.1 & 65.4 & 83.8 \\
1B    & 4M    & 50K   & 0.01 & 77.7 & 66.2 & 84.4 \\
\arrayrulecolor{lightgray}\midrule[0.25pt]\arrayrulecolor{black}
3B    & 20K   & 10K   & 0.03 & 55.6 & 43.3 & 62.5 \\
3B    & 30K   & 10K   & 0.03 & 61.4 & 49.2 & 68.7 \\
3B    & 60K   & 10K   & 0.03 & 68.1 & 55.5 & 75.7 \\
3B    & 120K  & 50K   & 0.03 & 71.2 & 58.6 & 78.7 \\
3B    & 400K  & 50K   & 0.03 & 75.0 & 62.8 & 82.1 \\
3B    & 1.2M  & 50K   & 0.03 & 76.4 & 64.7 & 83.4 \\
3B    & 2M    & 50K   & 0.03 & 76.9 & 64.7 & 83.7 \\
3B    & 4M    & 50K   & 0.01 & 77.6 & 65.6 & 84.3 \\
    \bottomrule
  \end{tabulary}
\end{table*}

\begin{table*}[h]
  \setlength{\tabcolsep}{5pt}
  \setlength{\extrarowheight}{5pt}
  \renewcommand{\arraystretch}{0.75}
  \centering
  \caption{Tabular representation of the finetune results (\%) for model \emph{ViT-s/16} on ImageNet, ImageNet V2 test set and ImageNet ReaL test set.}\label{tbl:xs_16_ft}
  \begin{tabulary}{1.0\textwidth}{l r c c c c c c c c}
    \toprule[1pt]
    \bf{Data Size} & \bf{Steps} & \bf{Cooldown} & \bf{LR} & \bf{ImageNet} & \bf{ImageNet V2} & \bf{ImageNet ReaL} \\
    \midrule
30M   & 20K   & 10K   & 0.03 & 56.0 & 43.2 & 63.0 \\
30M   & 30K   & 10K   & 0.03 & 62.2 & 49.4 & 69.5 \\
30M   & 60K   & 10K   & 0.03 & 67.8 & 54.8 & 75.3 \\
30M   & 120K  & 50K   & 0.03 & 70.0 & 57.5 & 77.7 \\
30M   & 400K  & 50K   & 0.03 & 73.7 & 60.9 & 81.0 \\
30M   & 1.2M  & 50K   & 0.03 & 75.0 & 62.4 & 82.0 \\
30M   & 2M    & 50K   & 0.01 & 75.2 & 63.0 & 82.3 \\
\arrayrulecolor{lightgray}\midrule[0.25pt]\arrayrulecolor{black}
300M  & 20K   & 10K   & 0.03 & 56.3 & 43.2 & 63.3 \\
300M  & 30K   & 10K   & 0.03 & 62.0 & 49.5 & 69.4 \\
300M  & 60K   & 10K   & 0.03 & 67.4 & 54.3 & 75.0 \\
300M  & 120K  & 50K   & 0.03 & 70.1 & 57.8 & 77.6 \\
300M  & 400K  & 50K   & 0.03 & 73.6 & 61.2 & 80.6 \\
300M  & 1.2M  & 50K   & 0.03 & 74.9 & 62.8 & 82.0 \\
300M  & 2M    & 50K   & 0.01 & 75.4 & 63.4 & 82.6 \\
\arrayrulecolor{lightgray}\midrule[0.25pt]\arrayrulecolor{black}
1B    & 20K   & 10K   & 0.03 & 56.2 & 44.1 & 63.2 \\
1B    & 30K   & 10K   & 0.03 & 62.4 & 49.7 & 69.8 \\
1B    & 60K   & 10K   & 0.03 & 68.0 & 54.9 & 75.6 \\
1B    & 120K  & 50K   & 0.03 & 70.5 & 57.5 & 77.8 \\
1B    & 400K  & 50K   & 0.03 & 73.9 & 61.6 & 81.1 \\
1B    & 1.2M  & 50K   & 0.03 & 75.1 & 63.2 & 82.1 \\
\arrayrulecolor{lightgray}\midrule[0.25pt]\arrayrulecolor{black}
3B    & 20K   & 10K   & 0.03 & 56.4 & 43.6 & 63.3 \\
3B    & 30K   & 10K   & 0.03 & 62.6 & 49.9 & 70.1 \\
    \bottomrule
  \end{tabulary}
\end{table*}

\begin{table*}[h]
  \setlength{\tabcolsep}{5pt}
  \setlength{\extrarowheight}{5pt}
  \renewcommand{\arraystretch}{0.75}
  \centering
  \caption{Tabular representation of the finetune results (\%) for model \emph{ViT-S/32} on ImageNet, ImageNet V2 test set and ImageNet ReaL test set.}\label{tbl:s_32_ft}
  \begin{tabulary}{1.0\textwidth}{l r c c c c c c c c}
    \toprule[1pt]
    \bf{Data Size} & \bf{Steps} & \bf{Cooldown} & \bf{LR} & \bf{ImageNet} & \bf{ImageNet V2} & \bf{ImageNet ReaL} \\
    \midrule
30M   & 20K   & 10K   & 0.03 & 59.3 & 47.1 & 66.3 \\
30M   & 30K   & 10K   & 0.03 & 64.3 & 51.8 & 71.7 \\
30M   & 60K   & 10K   & 0.03 & 70.3 & 58.1 & 77.5 \\
30M   & 120K  & 50K   & 0.03 & 73.4 & 61.2 & 80.5 \\
30M   & 400K  & 50K   & 0.03 & 77.1 & 65.7 & 83.6 \\
30M   & 1.2M  & 50K   & 0.03 & 79.0 & 67.3 & 84.9 \\
30M   & 2M    & 50K   & 0.03 & 79.1 & 67.9 & 85.1 \\
30M   & 4M    & 50K   & 0.01 & 79.7 & 68.2 & 85.6 \\
\arrayrulecolor{lightgray}\midrule[0.25pt]\arrayrulecolor{black}
300M  & 20K   & 10K   & 0.03 & 59.3 & 47.1 & 66.2 \\
300M  & 30K   & 10K   & 0.03 & 64.2 & 51.0 & 71.5 \\
300M  & 60K   & 10K   & 0.03 & 70.1 & 57.6 & 77.4 \\
300M  & 120K  & 50K   & 0.03 & 73.4 & 60.5 & 80.6 \\
300M  & 400K  & 50K   & 0.03 & 77.5 & 66.3 & 84.0 \\
300M  & 1.2M  & 50K   & 0.03 & 79.0 & 67.9 & 85.1 \\
300M  & 2M    & 50K   & 0.03 & 79.6 & 67.8 & 85.6 \\
300M  & 4M    & 50K   & 0.03 & 79.9 & 68.5 & 85.8 \\
\arrayrulecolor{lightgray}\midrule[0.25pt]\arrayrulecolor{black}
1B    & 20K   & 10K   & 0.03 & 59.0 & 46.2 & 66.2 \\
1B    & 30K   & 10K   & 0.03 & 64.0 & 51.4 & 71.4 \\
1B    & 60K   & 10K   & 0.03 & 70.5 & 57.7 & 77.7 \\
1B    & 120K  & 50K   & 0.03 & 73.6 & 60.8 & 80.7 \\
1B    & 400K  & 50K   & 0.03 & 77.6 & 65.7 & 84.0 \\
1B    & 1.2M  & 50K   & 0.03 & 79.5 & 68.0 & 85.5 \\
1B    & 2M    & 50K   & 0.03 & 79.7 & 68.2 & 85.5 \\
1B    & 4M    & 50K   & 0.03 & 80.2 & 68.1 & 85.9 \\
\arrayrulecolor{lightgray}\midrule[0.25pt]\arrayrulecolor{black}
3B    & 20K   & 10K   & 0.03 & 59.3 & 47.3 & 66.4 \\
3B    & 30K   & 10K   & 0.03 & 64.3 & 51.5 & 71.6 \\
3B    & 60K   & 10K   & 0.03 & 70.2 & 57.2 & 77.6 \\
3B    & 120K  & 50K   & 0.03 & 73.5 & 61.3 & 80.7 \\
3B    & 400K  & 50K   & 0.03 & 77.6 & 65.7 & 84.0 \\
3B    & 1.2M  & 50K   & 0.03 & 79.4 & 67.4 & 85.4 \\
3B    & 2M    & 50K   & 0.01 & 79.5 & 68.5 & 85.6 \\
3B    & 4M    & 50K   & 0.01 & 79.9 & 69.4 & 86.0 \\
    \bottomrule
  \end{tabulary}
\end{table*}

\begin{table*}[h]
  \setlength{\tabcolsep}{5pt}
  \setlength{\extrarowheight}{5pt}
  \renewcommand{\arraystretch}{0.75}
  \centering
  \caption{Tabular representation of the finetune results (\%) for model \emph{ViT-s/28} on ImageNet, ImageNet V2 test set and ImageNet ReaL test set.}\label{tbl:xs_28_ft}
  \begin{tabulary}{1.0\textwidth}{l r c c c c c c c c}
    \toprule[1pt]
    \bf{Data Size} & \bf{Steps} & \bf{Cooldown} & \bf{LR} & \bf{ImageNet} & \bf{ImageNet V2} & \bf{ImageNet ReaL} \\
    \midrule
30M   & 20K   & 10K   & 0.03 & 50.3 & 38.0 & 56.9 \\
30M   & 30K   & 10K   & 0.03 & 55.8 & 43.4 & 62.8 \\
30M   & 60K   & 10K   & 0.03 & 61.5 & 48.5 & 68.8 \\
30M   & 120K  & 50K   & 0.03 & 64.1 & 51.4 & 71.6 \\
30M   & 400K  & 50K   & 0.03 & 68.4 & 55.5 & 75.7 \\
30M   & 1.2M  & 50K   & 0.03 & 69.8 & 57.2 & 77.4 \\
30M   & 2M    & 50K   & 0.01 & 70.2 & 57.5 & 77.8 \\
\arrayrulecolor{lightgray}\midrule[0.25pt]\arrayrulecolor{black}
300M  & 20K   & 10K   & 0.03 & 50.3 & 38.2 & 56.9 \\
300M  & 30K   & 10K   & 0.03 & 55.7 & 43.6 & 62.7 \\
300M  & 60K   & 10K   & 0.03 & 61.1 & 48.8 & 68.5 \\
300M  & 120K  & 50K   & 0.03 & 64.0 & 51.1 & 71.5 \\
300M  & 400K  & 50K   & 0.03 & 68.6 & 55.5 & 76.0 \\
300M  & 1.2M  & 50K   & 0.03 & 70.1 & 57.1 & 77.6 \\
300M  & 2M    & 50K   & 0.03 & 70.5 & 57.1 & 77.9 \\
\arrayrulecolor{lightgray}\midrule[0.25pt]\arrayrulecolor{black}
1B    & 20K   & 10K   & 0.03 & 49.9 & 37.8 & 56.5 \\
1B    & 30K   & 10K   & 0.03 & 55.2 & 42.8 & 62.3 \\
1B    & 60K   & 10K   & 0.03 & 61.0 & 47.9 & 68.4 \\
1B    & 120K  & 50K   & 0.03 & 64.0 & 51.1 & 71.5 \\
1B    & 400K  & 50K   & 0.03 & 68.5 & 55.7 & 75.9 \\
1B    & 1.2M  & 50K   & 0.03 & 70.0 & 57.3 & 77.3 \\
\arrayrulecolor{lightgray}\midrule[0.25pt]\arrayrulecolor{black}
3B    & 20K   & 10K   & 0.03 & 49.9 & 38.0 & 56.3 \\
3B    & 30K   & 10K   & 0.03 & 55.4 & 43.5 & 62.4 \\
    \bottomrule
  \end{tabulary}
\end{table*}

\end{document}